\documentclass{article}
\PassOptionsToPackage{numbers,sort&compress}{natbib}
\usepackage{iclr2027_conference,times}
\setcitestyle{numbers,square,citesep={, }}
\usepackage[T1]{fontenc}
\usepackage{amsmath,amssymb,graphicx,booktabs,multirow,array,tabularx}
\usepackage{ragged2e,enumitem,subcaption,placeins,float,flafter}
\usepackage{pdflscape}
\usepackage{needspace}
\usepackage[breakable]{tcolorbox}
\usepackage{microtype}
\usepackage[hidelinks]{hyperref}
\usepackage{url}
\graphicspath{{figures/}}
\hypersetup{
  pdftitle={Detecting Time Series Anomalies Like an Expert: A Multi-Agent LLM Framework with Specialized Analyzers},
  pdfauthor={Hyeongwon Kang, Jeongseob Kim, Jinwoo Park, Pilsung Kang}
}
\newtcolorbox{promptbox}[1][]{breakable,colback=gray!4,colframe=gray!60,title={#1},fonttitle=\bfseries\small,fontupper=\footnotesize,boxrule=0.5pt,arc=1pt,left=5pt,right=5pt,top=4pt,bottom=4pt}
\newtcolorbox{agentbox}[1][]{breakable,colback=blue!3,colframe=blue!45,title={#1},fonttitle=\bfseries\small,fontupper=\footnotesize,boxrule=0.5pt,arc=1pt,left=5pt,right=5pt,top=4pt,bottom=4pt}
\newtcolorbox{detectorbox}[1][]{breakable,colback=red!3,colframe=red!45,title={#1},fonttitle=\bfseries\small,fontupper=\footnotesize,boxrule=0.5pt,arc=1pt,left=5pt,right=5pt,top=4pt,bottom=4pt}
\newtcolorbox{supervisorbox}[1][]{breakable,colback=gray!4,colframe=gray!60,title={#1},fonttitle=\bfseries\small,fontupper=\footnotesize,boxrule=0.5pt,arc=1pt,left=5pt,right=5pt,top=4pt,bottom=4pt}

\title{Detecting Time Series Anomalies\\Like an Expert: A Multi-Agent LLM\\Framework with Specialized Analyzers}
\author{Hyeongwon Kang$^{1}$ \qquad Jeongseob Kim$^{2}$\\
Jinwoo Park$^{3}$ \qquad Pilsung Kang$^{3,*}$\\[0.5em]
\normalfont\small $^{1}$Department of Industrial and Management Engineering, Korea University\\
$^{2}$Mirae Asset Securities\\
$^{3}$Department of Industrial Engineering, Seoul National University\\[0.3em]
\texttt{hyeongwon\_kang@korea.ac.kr} \quad \texttt{jeongseob.kim@miraeasset.com}\\
\texttt{jinwoo\_park@snu.ac.kr} \quad \texttt{pilsung\_kang@snu.ac.kr}\\
$^{*}$Corresponding author}
\iclrfinalcopy
\begin{document}
\maketitle
\lhead{}
\begin{abstract}
Time-series anomaly detection often returns scores or intervals, while analysts need to understand the abnormal behavior and the evidence supporting it. We introduce SAGE (Specialized Analyzer Group for Expert-like Detection), a multi-agent framework for evidence-grounded diagnosis of univariate time series. Four specialized Analyzers examine point, structural, seasonal, and pattern anomalies using numerical tools and diagnostic visualizations. A Detector integrates their evidence into intervals, candidate types, and evidence-strength confidence scores; a Supervisor translates these records into analyst-facing reports. Synthetic in-context references are constructed from normal-reference training segments, reducing dependence on real anomalous demonstrations. Across Yahoo S5, KPI, and WSD, SAGE achieves an average Point-F1 of 66.26, the highest among the evaluated methods. Controlled synthetic evaluation examines localization and type diagnosis, while component ablations support the detection contribution of specialized evidence generation. In a method-blind human study, evaluators rate SAGE's diagnostic outputs as more useful than those of the compared methods.
\end{abstract}

\section{Introduction}
\label{sec:introduction}
Time-series anomaly detection supports monitoring of web services, infrastructure, and industrial processes \citep{DBLP:conf/kdd/LaptevAF15,DBLP:conf/kdd/RenXWYHKXYTZ19}. A detected interval is often only the start of an analyst's investigation. The analyst must distinguish an isolated spike from a persistent level shift, determine whether periodic structure has changed, and examine why a system regards the interval as abnormal. These questions require different evidence: local statistics for contextual outliers, segment comparisons for regime changes, and frequency or shape analysis for disrupted patterns. A single anomaly score does not directly communicate this diagnostic structure.

We therefore seek to organize heterogeneous numerical and visual evidence into an explicit diagnostic record. Here, \emph{diagnosis} means identifying abnormal temporal behavior, its location, plausible anomaly types, and supporting evidence. It does not mean establishing a causal root cause. A univariate signal may reveal a level shift without identifying the external event that caused it.

Large language and vision-language models offer a natural interface for evidence comparison and explanation. Recent systems use textual prompting, retrieved examples, or visual time-series representations for anomaly detection \citep{DBLP:journals/corr/abs-2405-14755,DBLP:conf/kdd/LiuZQMQBLR025,DBLP:journals/corr/abs-2411-02465}. However, direct reasoning over long numerical inputs remains an awkward fit for precise statistical operations and heterogeneous anomaly families. Tool use can provide the missing numerical precision, but a diagnostic system must also decide which evidence to generate, how to compare conflicting findings, and how to preserve that evidence in its final output.

We propose SAGE, a multi-agent framework that decomposes diagnosis by anomaly family. Four Analyzers specialize in point, structural, seasonal, and pattern anomalies. Each uses dedicated numerical tools and visual representations to produce candidate regions and diagnostic evidence. A Detector consolidates these findings into anomaly intervals, candidate types, and confidence scores interpreted as evidence strength. A Supervisor turns the structured records into reports describing the signal, alarm rationale, and suggested follow-up checks. The role of the LLM is to organize and interpret tool-supported evidence, while numerical tools retain access to the original signal.

Real anomaly examples and anomaly-type annotations are often scarce, especially early in deployment. The Detector nevertheless benefits from examples that contrast expected behavior with a specific deviation when interpreting heterogeneous Analyzer evidence. SAGE therefore constructs synthetic in-context learning (ICL) references from normal-reference training segments. Rule-based perturbations create matched normal--anomalous pairs, and retrieval supplies contrasts relevant to the Analyzer-proposed types. Because each pair shares the same underlying normal pattern, the contrast highlights the injected change rather than unrelated differences between series. This provides type-specific reference evidence without relying on real anomalous demonstrations.

Our contributions are as follows:
\begin{enumerate}[nosep,leftmargin=*]
  \item We formulate univariate anomaly diagnosis as structured evidence generation and aggregation, linking family-specific analysis to explicit records of anomaly intervals, candidate types, and supporting evidence.
  \item We develop a multi-agent workflow that combines specialized Analyzers, numerical and visual tools, compact prompt representations, and synthetic contrastive references constructed from normal training segments.
  \item We evaluate detection, type diagnosis, component contributions, backbone dependence, and human-rated report quality, showing the value of evidence-grounded organization across complementary assessments.
\end{enumerate}

\section{Related Work}
\label{sec:related}
\paragraph{Time-series anomaly detection.}
Classical methods identify distributional or neighborhood deviations, as in Isolation Forest and LOF \citep{liu2008isolation,DBLP:conf/sigmod/BreunigKNS00}. Neural methods use reconstruction, latent dynamics, or attention-based discrepancies, including LSTM-AE, USAD, OmniAnomaly, and Anomaly Transformer \citep{DBLP:journals/corr/MalhotraRAVAS16,DBLP:conf/kdd/AudibertMGMZ20,DBLP:conf/kdd/SuZNLSP19,DBLP:conf/iclr/XuWWL22}. TFAD further uses decomposition and time--frequency analysis to model temporal structure \citep{DBLP:conf/cikm/ZhangZWS22}. These methods provide strong detection mechanisms, but a score alone does not specify whether a deviation is point-like, structural, seasonal, or pattern-based, or record the evidence for that assignment. SAGE uses established statistical and signal-processing tools for this diagnostic purpose rather than proposing a new low-level detector.

\paragraph{LLM/VLM-based anomaly analysis.}
PromptCast, LLMTime, and Time-LLM show how language models can be adapted to time-series forecasting \citep{DBLP:journals/tkde/XueS24,DBLP:conf/nips/GruverFQW23,DBLP:conf/iclr/0005WMCZSCLLPW24}; forecasting does not directly provide anomaly intervals or type-specific diagnostic evidence.
SigLLM evaluates language-model prompting and forecasting pipelines for time-series anomalies \citep{DBLP:journals/corr/abs-2405-14755}. LLMAD combines anomaly-oriented reasoning with retrieved real examples \citep{DBLP:conf/kdd/LiuZQMQBLR025}. TAMA uses visual representations and few-shot multimodal reasoning \citep{DBLP:journals/corr/abs-2411-02465}, while TSAD-Agents explores multimodal agentic analysis \citep{DBLP:conf/www/XuWLYZS26}. SAGE combines family-specific evidence generation with explicit Detector-level aggregation. Its synthetic references pair normal prototypes with controlled perturbations, and its output separates evidence-strength scoring, type diagnosis, and report generation.

\paragraph{Tool-supported and multi-agent reasoning.}
ReAct and Toolformer demonstrate how external tools can support language-model reasoning \citep{DBLP:conf/iclr/YaoZYDSN023,DBLP:conf/nips/SchickDDRLHZCS23}; MetaGPT and CAMEL study collaboration through specialized roles \citep{DBLP:conf/iclr/HongZCZCWZWYLZR24,DBLP:conf/nips/LiHIKG23}. SAGE applies these principles to a domain-specific decomposition whose roles follow the evidence required by anomaly families rather than generic conversational tasks. The Analyzers express candidate locations, types, and tool outputs in a shared representation; the Detector weighs supporting and conflicting evidence with an explicit rubric rather than treating agent findings as independent votes. The contribution is this diagnostic orchestration of existing tools and model capabilities, not a new numerical detector or foundation-model architecture.

\section{SAGE: Evidence-Grounded Anomaly Diagnosis}
\label{sec:method}
\subsection{Diagnostic outputs and architecture}
For a univariate sequence $X=(x_1,\ldots,x_T)$, SAGE produces
\begin{equation}
\mathcal{Y}=\{(s_j,e_j,c_j,\tau_j,d_j)\}_{j=1}^{m},
\label{eq:records}
\end{equation}
where $[s_j,e_j]$ is an anomaly interval, $c_j\in[0,1]$ is an evidence-strength score, $\tau_j\subseteq\mathcal{L}$ contains candidate anomaly types, and $d_j$ records diagnostic evidence and explanation. The taxonomy $\mathcal{L}$ contains nine types in four families: global and contextual point anomalies; trend, mean, and variance changes; amplitude and seasonality changes; and pattern shifts and waveform distortions. These types describe observable temporal behavior rather than external causes. The families organize complementary evidence rather than mutually exclusive detection gates: several Analyzers can support the same interval, and a record can contain multiple candidate types. This does not make the taxonomy exhaustive or adaptive. Definitions and injection rules appear in Appendix~\ref{app:taxonomy}.

Figure~\ref{fig:architecture} shows the pipeline. The original signal remains available to numerical tools, while LLM prompts receive a compact representation. Four Analyzers operate in parallel and pass their findings to the Detector. Retrieved synthetic references supplement this aggregation stage. Finally, the Supervisor converts Detector-selected records into a report. This separation makes the source of an observation, its interpretation, and its final presentation explicit.

\begin{figure}[H]
\centering
\includegraphics[width=\linewidth]{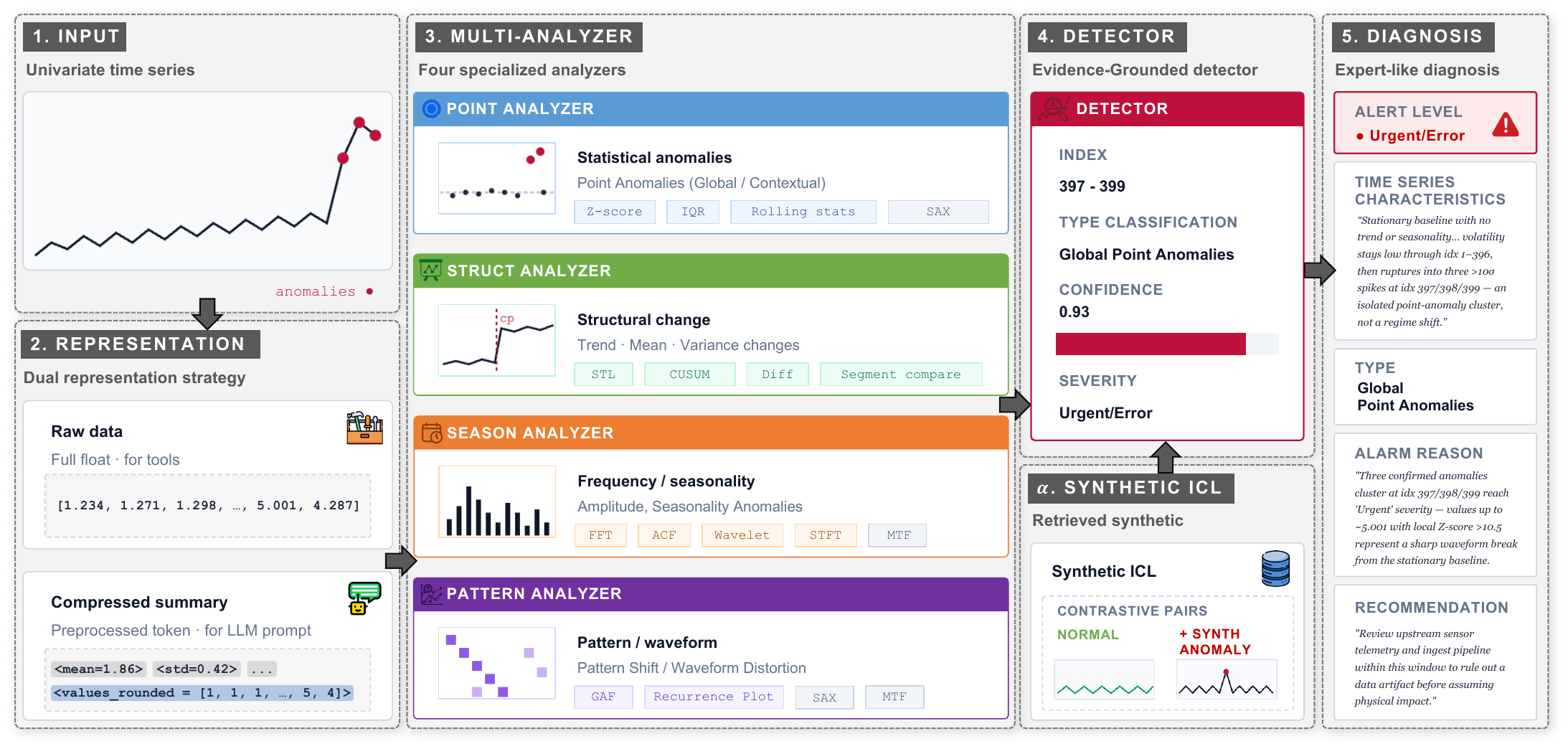}
\caption{SAGE decomposes univariate anomaly diagnosis into dual representations, four specialized Analyzers, evidence aggregation, and an analyst-facing report. Synthetic references support the Detector.}
\label{fig:architecture}
\end{figure}

\subsection{Dual representations and specialized evidence generation}
Numerical tools require the original floating-point values; language-model prompts require a compact description. SAGE therefore preserves $X$ for computation and constructs $\widetilde X$ from global statistics, representative integer-rounded samples, segment summaries, and extrema-preserving indices. In the reported implementation, compression reduces the numerical input token usage by approximately 75\% for a 400-point window. This saving concerns the input representation, not the total token cost of the multi-agent workflow.

Each Analyzer maps the two representations to a family-specific result:
\begin{equation}
O_r=A_r(X,\widetilde X),\qquad r\in\{\mathrm{pt},\mathrm{str},\mathrm{sea},\mathrm{pat}\}.
\label{eq:analyzers}
\end{equation}
\textbf{PointAnalyzer} uses global statistics, Z-scores, interquartile ranges, and multi-scale rolling statistics to identify global and contextual point deviations. \textbf{StructAnalyzer} examines persistent changes through decomposition, differences, change-point tests, and segment comparisons. Its tools include STL-style decomposition and CUSUM-based change-point analysis \citep{cleveland1990stl,page1954cusum}. \textbf{SeasonAnalyzer} examines oscillation amplitude and periodic organization through autocorrelation, Fourier transforms, wavelets, and short-time Fourier analysis. \textbf{PatternAnalyzer} examines motifs and waveform shape using symbolic aggregate approximation, recurrence plots, and visual encodings \citep{DBLP:conf/dmkd/LinKLC03,eckmann1987recurrence,wang2015gaf}.

Analyzers may use shared auxiliary tools when complementary evidence is useful. Their internal outputs differ---for example, outlier indices, change points, spectral summaries, or recurrence diagnostics---but the Detector receives a common representation of candidate locations, candidate types, analysis summaries, and supporting tool outputs. Specialization thus constrains the question each agent asks without requiring all tools to return the same numerical statistic. Appendix~\ref{app:tools} maps tools to roles, and Appendix~\ref{app:prompts} gives the agent prompts.

\subsection{Evidence aggregation and diagnostic reporting}
The Detector combines Analyzer outputs $\mathcal{O}=\{O_r\}_r$, diagnostic visualizations $\mathcal{V}$, and retrieved references $\mathcal{R}$:
\begin{equation}
\mathcal{C}=D(\mathcal{O},\mathcal{V},\mathcal{R}).
\label{eq:detector}
\end{equation}
Here, $\mathcal{C}=\{(s_j,e_j,c_j,\tau_j)\}_{j=1}^{m}$ contains the Detector-selected records. The Detector assigns integer scores from 0 to 100, normalized to $[0,1]$, using a prompt-defined rubric. The rubric considers deviation magnitude, statistical evidence, consistency across sources, and conflicting observations. Weak or ambiguous findings receive low scores; strong deviations or corroborating sources receive high scores. The implementation suppresses records below 50 on the original scale. Agreement can strengthen a candidate, but confidence is not simply a count of agreeing agents, nor is it a calibrated probability.

When Analyzers disagree, the Detector weighs the underlying tool and visual evidence rather than voting over agent outputs. A weak, isolated deviation can receive less support than a candidate with a clear magnitude or persistent change corroborated by another source. Retaining the interval and candidate types makes this judgment inspectable.

The Detector uses tool-reported indices to preserve localization precision and visual evidence to assess the extent of a candidate. Its prompt also permits reporting a missed candidate when the provided data or chart supports it. Thus, aggregation is evidence-guided but is not a formal constraint guaranteeing that every output is a member of the Analyzer candidate set. The prompts encode the intended behavior; they do not establish factual correctness of the generated interpretation.

The Supervisor combines $\mathcal{C}$ with Analyzer evidence to produce the diagnostic descriptions $d_j$ in $\mathcal{Y}$. It presents an executive summary, time-series characteristics, anomaly details, alarm level, alarm rationale, and follow-up recommendations. It is instructed not to introduce new intervals, types, or unsupported causes. Recommendations guide further investigation rather than certify an external cause.

\subsection{Synthetic in-context references from normal data}
\begin{figure}[t]
\centering
\includegraphics[width=\linewidth]{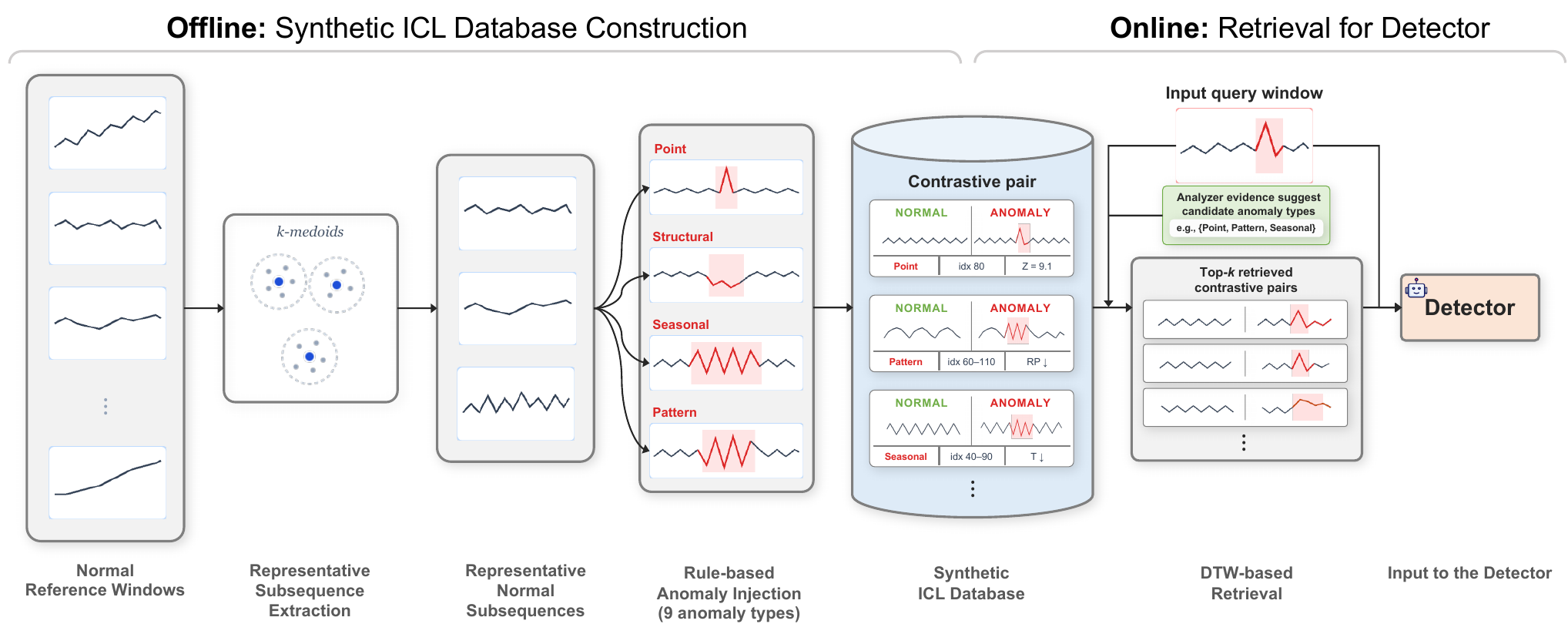}
\caption{Offline construction and online retrieval of synthetic contrastive references. DTW retrieves normal prototypes; Analyzer-proposed types determine which associated perturbations are passed to the Detector.}
\label{fig:icl}
\end{figure}
Real anomaly examples can be scarce or poorly annotated. SAGE builds references from the training portion of each time series, excluding segments that overlap labeled anomalies. It selects representative normal subsequences with $k$-medoids, retaining observed prototypes rather than averaged centroids. Each prototype receives one rule-based perturbation for each of the nine anomaly types. These injection rules construct references; they are not applied as inference-time detection rules. The original and perturbed signals form a contrastive pair with an injected type label and a summary of changed tool evidence (Figure~\ref{fig:icl}).

At inference time, Dynamic Time Warping with LB\_Keogh pruning retrieves the three closest normal prototypes \citep{sakoe1978dtw,keogh2002lbkeogh}. Analyzer-proposed candidate types select the associated synthetic contrasts. These references are supplied only to the Detector to support evidence comparison, confidence scoring, and type ranking. Neither retrieval nor type selection uses test labels or true test anomaly types. During reference construction, binary training labels are used only to exclude anomalous segments; real anomalous segments are not supplied as demonstrations.

The synthetic type evaluation is separate: it disables synthetic ICL for SAGE, so its test cases receive no retrieved synthetic demonstrations. This evaluates the existing tools and aggregation process on controlled perturbations, rather than reference-label copying. The taxonomy and perturbation design nevertheless define the scope of the evaluation; success does not establish coverage of all real failure modes.

\FloatBarrier
\section{Experimental Evaluation}
\label{sec:experiments}
\subsection{Setup and evaluation protocol}
We evaluate Yahoo S5 (367 series, average length approximately 1,400), KPI (29 series, approximately 100,000), and WSD (111 series, approximately 17,000) \citep{yahoo_webscope_s5,li2022constructing,zhang2022efficient}. Yahoo contains real web traffic (A1) and synthetic subsets (A2--A4); KPI and WSD contain operational infrastructure and web-service metrics. All three provide binary anomaly labels. Training and test data preserve temporal order. Yahoo uses 40\% of each series for training and 60\% for testing; KPI and WSD use 70\% for training and 5\% for testing, excluding the remaining 25\% to limit API inference time and cost. Their reported results therefore concern these reduced test portions. SAGE uses the training portion for normal-reference construction, not parameter learning.

The LangGraph workflow uses non-overlapping 400-point windows, temperature 0.0, and retrieval $k=3$. The common backbone for SAGE and the LLM/VLM comparisons is Claude Sonnet 4.6. Baseline input representations, prompting logic, ICL strategies, and post-processing are retained as closely as possible when replacing the backbone. This controls model identity, but does not equalize inference budgets. We compare eight ML/DL methods and four LLM/VLM systems. Additional settings and backbone experiments are given in Appendices~\ref{app:implementation} and~\ref{app:additional_results}.

We report Point-F1, point-adjusted F1 (PA-F1), Affiliation-F1 \citep{DBLP:conf/kdd/HuetNR22}, and Delayed-F1 with delay parameter $k=3$, all as percentages. Point-F1 computes precision and recall at the time-point level and penalizes imprecise boundaries and false positives. PA-F1 gives segment-level credit when a predicted anomaly overlaps a labeled anomaly, providing a more relaxed view of detection. Affiliation-F1 measures the temporal proximity of predicted and labeled anomaly intervals. Delayed-F1 assesses timely detection by treating an anomaly as detected when its first prediction occurs within three time steps after onset. Together, these metrics distinguish point-level localization, segment coverage, temporal proximity, and alert delay.

\subsection{Detection performance}
\begin{table}[!ht]
\centering
\footnotesize
\setlength{\tabcolsep}{1.6pt}
\renewcommand{\arraystretch}{1.0}
\caption{Main detection results (percent) for Yahoo S5, KPI, WSD, and their unweighted average. Pt, PA, Aff, and Del denote Point-F1, PA-F1, Affiliation-F1, and Delayed-F1, respectively; column maxima are bold.}
\label{tab:main}
\resizebox{\textwidth}{!}{%
\begin{tabular}{@{}l|cccc|cccc|cccc|cccc@{}}
\toprule
\multirow{2}{*}{Method}
& \multicolumn{4}{c}{Yahoo S5}
& \multicolumn{4}{c}{KPI}
& \multicolumn{4}{c}{WSD}
& \multicolumn{4}{c}{Average} \\
\cmidrule(lr){2-5} \cmidrule(lr){6-9} \cmidrule(lr){10-13} \cmidrule(lr){14-17}
& Pt & PA & Aff & Del
& Pt & PA & Aff & Del
& Pt & PA & Aff & Del
& Pt & PA & Aff & Del \\
\midrule
iForest   & 51.35 & 51.97 & 69.25 & 53.95 & 39.89 & 77.87 & 90.03 & 49.76 & 33.13 & 79.12 & 84.08 & 55.02 & 41.46 & 69.65 & 81.12 & 52.91 \\
OCSVM     & 50.55 & 52.23 & 67.54 & 53.28 & 39.86 & 75.81 & 89.58 & 53.86 & 34.97 & 87.61 & 91.56 & 74.07 & 41.79 & 71.88 & 82.90 & 60.40 \\
LOF       & 46.10 & 47.36 & 67.32 & 47.81 & 6.59 & 13.78 & 68.90 & 10.71 & 10.39 & 26.20 & 72.79 & 28.21 & 21.03 & 29.12 & 69.67 & 28.91 \\
\midrule
LSTM-AE   & 54.69 & 55.71 & 83.04 & 57.38 & 27.76 & 69.58 & 84.41 & 26.61 & 47.14 & 80.77 & 89.15 & 54.63 & 43.20 & 68.69 & 85.53 & 46.21 \\
LSTM-VAE  & 51.77 & 52.60 & 68.09 & 54.85 & 42.22 & 74.90 & \textbf{90.13} & 52.55 & 44.24 & 91.95 & 89.18 & 76.51 & 46.08 & 73.15 & 82.47 & 61.30 \\
USAD      & 49.11 & 50.40 & 67.39 & 52.45 & 30.71 & 65.91 & 86.63 & 52.87 & 32.44 & 78.06 & 85.78 & 60.59 & 37.42 & 64.79 & 79.93 & 55.31 \\
OmniAnom  & 45.65 & 49.71 & 75.55 & 51.00 & 14.83 & 44.13 & 72.17 & 13.65 & 39.55 & 83.03 & 81.77 & 44.73 & 33.34 & 58.96 & 76.50 & 36.46 \\
AnomalyTF & 13.27 & 16.85 & 52.72 & 49.70 & 38.38 & 58.62 & 84.81 & 55.67 & 42.34 & 60.04 & 91.16 & 87.06 & 31.33 & 45.17 & 76.23 & 64.14 \\
\midrule
SigLLM    & 53.95 & 55.02 & 86.44 & 58.09 & 11.72 & 27.30 & 68.85 & 4.61 & 28.72 & 58.98 & 69.64 & 19.51 & 31.46 & 47.10 & 74.98 & 27.40 \\
LLMAD     & 76.82 & 77.12 & \textbf{88.93} & 83.38 & 19.47 & 44.21 & 84.66 & 48.02 & 41.59 & 71.54 & 79.36 & 77.04 & 45.96 & 64.29 & 84.32 & 69.48 \\
TAMA      & 3.82 & 3.95 & 67.35 & 61.09 & 4.89 & 7.48 & 67.51 & 34.16 & 14.79 & 16.55 & 64.96 & 62.41 & 7.84 & 9.33 & 66.61 & 52.55 \\
TSAD-Ag   & 32.01 & 32.25 & 61.64 & 39.03 & 15.09 & 25.09 & 52.97 & 9.01 & 22.89 & 24.92 & 37.56 & 30.37 & 23.33 & 27.42 & 50.72 & 26.14 \\
\midrule
\textbf{SAGE} & \textbf{79.69} & \textbf{80.46} & 88.37 & \textbf{84.32} & \textbf{60.97} & \textbf{78.61} & 86.42 & \textbf{74.83} & \textbf{58.12} & \textbf{95.68} & \textbf{96.82} & \textbf{97.08} & \textbf{66.26} & \textbf{84.92} & \textbf{90.54} & \textbf{85.41} \\
\bottomrule
\end{tabular}
}
\end{table}

Table~\ref{tab:main} reports all four metrics for each dataset and their unweighted averages. SAGE achieves 66.26 Point-F1, 84.92 PA-F1, 90.54 Affiliation-F1, and 85.41 Delayed-F1. Its average Point-F1 improves by 43.8\% relative to LSTM-VAE and 44.2\% relative to LLMAD. Relative to LLMAD, the absolute Point-F1 gains are larger on KPI and WSD than on Yahoo: SAGE obtains 60.97 and 58.12 Point-F1 compared with 19.47 and 41.59 for LLMAD.

The gains vary with signal characteristics. On Yahoo's real A1 subset, SAGE obtains 75.31 Point-F1 versus LLMAD's 54.03. LLMAD remains stronger on A2 and A3, while SAGE has a small advantage on A4. SAGE is also not best on every metric: LLMAD has higher Yahoo Affiliation-F1, and LSTM-VAE has higher KPI Affiliation-F1. These results support broad average improvements rather than uniform superiority. Appendix~\ref{app:additional_results} presents the Yahoo subset results and further component and backbone analyses.

\paragraph{Lightweight alternatives.}
A supplementary comparison evaluates Isolation Forest, Matrix Profile, STL residual analysis, and their average-score ensemble (Appendix~\ref{app:lightweight}). In that recorded comparison, SAGE reaches 66.3 average Point-F1, versus 53.2 for STL residuals and 43.1 for the ensemble. This supports an advantage over the evaluated alternatives, but the ensemble's weaker performance than STL alone limits conclusions about lightweight fusion generally. The comparison uses different evidence generators from SAGE and therefore does not isolate the LLM's contribution on identical tool outputs.

\subsection{Controlled anomaly-type diagnosis}
\begin{table}[H]
\centering
\setlength{\tabcolsep}{4pt}
\renewcommand{\arraystretch}{0.95}
\caption{Synthetic evaluation with SAGE's synthetic ICL disabled (percent). Recall measures family-specific Analyzer evidence coverage; type-agreement rates are reported summaries with unspecified aggregation (Appendix~\ref{app:type_aggregation}). Baseline columns measure final localization.}
\label{tab:types}
\begin{subtable}[t]{0.48\textwidth}
\centering
\footnotesize
\caption{Category-wise type evaluation}
\label{tab:types_family}
\begin{tabular}{@{}lrcc@{}}
\toprule
Category & N & Det. Rec. & Type Agr. \\
\midrule
Point & 40 & 90\% & 88\%\\
Structural & 60 & 82\% & \textbf{93\%}\\
Seasonal & 24 & 54\% & 75\%\\
Pattern & 32 & \textbf{97\%} & 31\%\\
\midrule
Overall & 156 & 83\% & 76\%\\
\bottomrule
\end{tabular}
\end{subtable}
\hfill
\begin{subtable}[t]{0.48\textwidth}
\centering
\footnotesize
\caption{SAGE vs. LLM/VLM baselines}
\label{tab:types_baselines}
\begin{tabular}{@{}lcccc@{}}
\toprule
Method & Pt-F1 & PA-F1 & Del-F1 & Aff-F1 \\
\midrule
SigLLM & 30.28 & 48.59 & 41.18 & 51.87 \\
LLMAD & 67.38 & 87.02 & 89.08 & 89.27 \\
TAMA & 65.69 & 72.31 & \textbf{93.08} & \textbf{93.71} \\
TSAD-Agents & 22.68 & 47.02 & 44.08 & 53.12 \\
\textbf{SAGE} & \textbf{72.34} & \textbf{87.93} & 89.57 & 91.03 \\
\bottomrule
\end{tabular}
\end{subtable}
\end{table}

Binary benchmark labels cannot establish type-diagnosis quality. We therefore evaluate 156 synthetic samples spanning the nine injected types, with synthetic ICL disabled for SAGE. Detection recall is the fraction for which the Analyzer assigned to the injected family returns an evidence interval overlapping the labeled anomaly. The reported type-agreement metric compares the highest-confidence predicted type with the injected type on detected cases. Its within-family and overall aggregation are not specified, so we treat the reported rates as descriptive summaries (Appendix~\ref{app:type_aggregation}). This recall concerns \emph{Analyzer evidence coverage}, not final Detector-level recall. Baselines are compared using localization metrics because their type vocabularies are not all aligned with ours.

Table~\ref{tab:types} separates these quantities. Point anomalies show 90\% evidence recall and 88\% reported type agreement; structural anomalies have the highest type agreement (93\%). Pattern anomalies are frequently located (97\% evidence recall), but have the lowest reported type agreement (31\%). Seasonal evidence recall is 54\%. These differences show that finding an abnormal region and assigning its injected type are distinct tasks, especially for pattern changes.

On the same synthetic set, SAGE has the strongest Point-F1 and PA-F1, whereas TAMA leads on Delayed-F1 and Affiliation-F1. TAMA's higher Delayed-F1 and Affiliation-F1 indicate that its predictions often occur soon after anomaly onset and near the labeled intervals, respectively. This is consistent with visual representations capturing approximate anomaly extent, even though its lower Point-F1 and PA-F1 indicate less precise localization. The aggregate comparison does not isolate visual input as the cause. An auxiliary, single-author annotation covers 300 benchmark test intervals, of which 251 are point anomalies. Its aggregate and per-type agreement summaries do not reconcile as a pooled proportion; we retain them only as supplementary records (Appendix~\ref{app:annotation}).

\section{Analysis}
\label{sec:analysis}
\subsection{Contributions of tools, specialization, vision, and references}
\begin{figure}[!htbp]
\centering
\includegraphics[width=\linewidth]{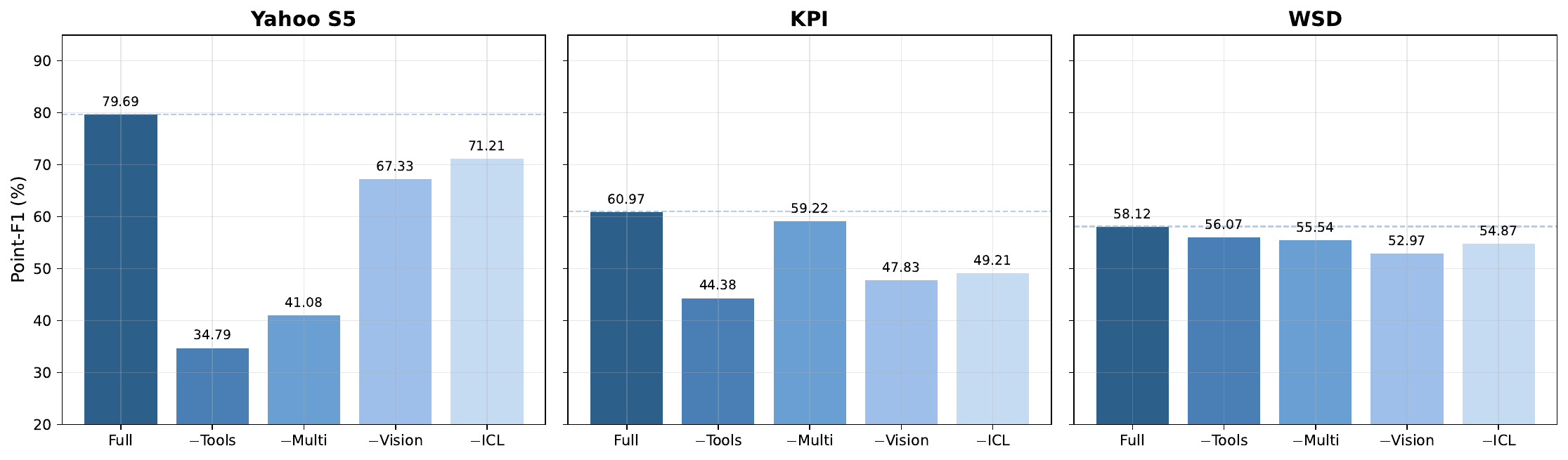}
\caption{Point-F1 component ablations on Yahoo S5, KPI, and WSD. Bars show the full system and the removal of tools, Analyzer specialization (Multi), visual evidence, or synthetic in-context references. Detailed results for all metrics are in Appendix~\ref{app:additional_results}.}
\label{fig:ablation}
\end{figure}
Figure~\ref{fig:ablation} shows the Point-F1 component ablations. On Yahoo, removing numerical tools reduces Point-F1 from 79.69 to 34.79, and removing Analyzer specialization reduces it to 41.08. These drops support the value of numerical evidence and family-specific analysis for heterogeneous anomalies. On KPI, removing visual evidence lowers Point-F1 from 60.97 to 47.83, and removing ICL lowers it to 49.21. The corresponding WSD scores fall from 58.12 to 52.97 and 54.87.

Specialization has a smaller Point-F1 effect on KPI and WSD. Some segment metrics even improve when it is removed: for example, WSD PA-F1 rises from 95.68 to 98.87. These observations qualify the contribution. Specialized analysis improves point-level localization in these experiments, but does not uniformly improve every overlap-based metric. The ablations also alter the workflow and its computational budget; they are not a matched-cost test of agent count alone. Removing tools establishes their importance, but does not isolate the benefit of LLM aggregation over a non-LLM aggregator receiving the same evidence.

Without ICL, SAGE still exceeds LLMAD with ICL on KPI (49.21 versus 19.47 Point-F1) and WSD (54.87 versus 41.59). On Yahoo, however, zero-shot SAGE scores 71.21 versus 76.28 for zero-shot LLMAD. Retrieved references and architecture therefore contribute differently across datasets. Removing ICL tests dependence on retrieved references; it does not test unseen anomaly types or robustness to misleading references. Full ablation and zero-shot tables appear in Appendix~\ref{app:additional_results}.

\subsection{Human evaluation of diagnostic outputs}
\begin{figure}[!htbp]
\centering
\includegraphics[width=0.82\textwidth]{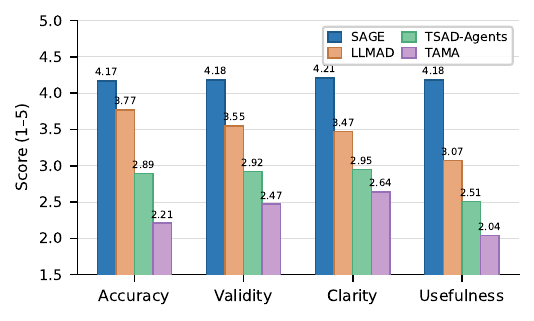}
\caption{Mean human ratings by criterion on a 1--5 scale (10 evaluators, 25 samples). Appendix~\ref{app:human} gives the rubric and evaluator-level scores.}
\label{fig:human}
\end{figure}
A method-blind study uses 10 technically experienced evaluators, 25 samples, and four methods. The sample set contains eight Yahoo, four KPI, four WSD, and nine synthetic examples, with one synthetic example per anomaly type. Evaluators view the input, ground-truth and predicted intervals, and diagnostic explanation. Method identities are anonymized and presentation order is randomized. Scores range from 1 to 5 for Accuracy, Validity, Clarity, and Usefulness; Appendix~\ref{app:human} contains the rubric and evaluator-level results.

SAGE receives the highest mean score for all four criteria (Figure~\ref{fig:human}), with an overall average of 4.18. Clarity is its highest-rated criterion (4.21); Usefulness is 4.18 versus 3.07 for LLMAD, whose Accuracy rating is 3.77. SAGE has the highest overall mean for nine of the ten evaluators and ties LLMAD for the tenth (Appendix~\ref{app:human}). The reported paired Wilcoxon tests yield $p<0.001$ against each baseline, but do not account for dependence among repeated evaluators and samples. We therefore emphasize the descriptive ratings rather than a dependence-aware significance claim. The study evaluates perceived diagnostic quality on these cases, not verified causal correctness or measured operational outcomes.

\section{Discussion and Conclusion}
\label{sec:conclusion}
SAGE organizes univariate anomaly analysis around specialized evidence generation, Detector-level integration, and structured reporting. The experiments show gains in average detection performance and higher human ratings of diagnostic outputs. Controlled type evaluation reveals that finding an abnormal region and assigning its type remain distinct challenges. The framework is best understood as evidence-grounded support for analysts, with confidence representing evidence strength and recommendations identifying follow-up checks.

Our conclusions are empirical and restricted to the evaluated univariate domains and anomaly settings. The framework remains limited by inference cost, backbone dependence, and a fixed anomaly taxonomy. Extending diagnosis to multivariate data will require evidence about variable contributions and changing inter-variable relationships; reducing all variables to one score can obscure this information. Future work should improve seasonal localization and pattern-type discrimination and study human-in-the-loop monitoring with variable-aware evidence. Causal root-cause identification requires additional domain knowledge and evidence beyond the observed univariate signal.

\label{page:main_end}
\clearpage
\section*{AI Use Statement}
LLMs and VLMs are components of SAGE and of the evaluated language-model baselines. Their roles in tool-supported analysis, evidence aggregation, scoring, and report generation are described in Sections~\ref{sec:method}--\ref{sec:analysis} and Appendix~\ref{app:prompts}. Generative AI assisted research planning and methodological feedback, interpretation of existing results, manuscript drafting and revision, language editing, and LaTeX formatting. The empirical measurements reported here originate from the authors' experiments; this assistance did not generate new experimental measurements. Synthetic reference and evaluation sequences in the reported method are constructed using the rule-based perturbations in Appendix~\ref{app:taxonomy}.

\section*{Ethics Statement}
SAGE is intended as analyst decision support. Incorrect or overconfident reports may lead to unnecessary intervention or missed incidents; generated explanations should therefore be checked against evidence, and external causes should not be inferred from a univariate signal alone. The human study involved voluntary assessment of anonymized model outputs, and results are reported using de-identified evaluator indices. Formal IRB approval or equivalent institutional review was not obtained before the study. The evaluation used public benchmark and synthetic signals and did not collect sensitive or personally identifying participant information. These study procedures do not establish operational safety or causal validity of the reports.

\section*{Reproducibility Statement}
Section~\ref{sec:method} specifies the diagnostic workflow, and Section~\ref{sec:experiments} describes datasets, model comparisons, and metrics. Appendices~\ref{app:taxonomy}--\ref{app:prompts} document synthetic perturbations, numerical parameters, the recorded software environment, tool assignments, and prompts. Appendix~\ref{app:runtime} reports recorded runtime and token usage. Appendix~\ref{app:additional_results} distinguishes the updated score summary from the earlier full-metric backbone record, and Appendix~\ref{app:human} describes the human evaluation. Hosted-model reproducibility depends on continued access to the reported model versions. Temperature zero limits sampling variation but is not a guarantee of identical hosted-API outputs across repeated calls or service changes.

\bibliography{references}
\bibliographystyle{iclr2027_conference}
\clearpage
\appendix
\raggedbottom
\section{Anomaly Taxonomy and Synthetic References}
\label{app:taxonomy}
The nine types provide a vocabulary for observable temporal deviations. Adjacent types can co-occur in real signals; the synthetic labels identify the dominant injected change rather than a unique causal explanation. Tables~\ref{tab:ch5_anomaly_types_rules} and~\ref{tab:ch5_synthetic_injection_params} retain the definitions and parameterization used in the experiments.

\begin{table}[H]
\centering
\footnotesize
\setlength{\tabcolsep}{2.2pt}
\renewcommand{\arraystretch}{0.92}
\caption{Definitions and synthetic injection rules of the nine anomaly types used in SAGE.}
\label{tab:ch5_anomaly_types_rules}
\begin{tabularx}{\textwidth}{@{}c>{\RaggedRight\arraybackslash}p{0.18\textwidth}>{\RaggedRight\arraybackslash}p{0.11\textwidth}>{\RaggedRight\arraybackslash}X>{\RaggedRight\arraybackslash}X@{}}
\toprule
ID & Type name & Family & Definition & Synthetic injection rule \\
\midrule
1 & Global point anomaly & Point
  & A point that strongly deviates from the global distribution of the series
  & Insert a large isolated spike or drop at a random position. \\
2 & Contextual point anomaly & Point
  & A point that appears abnormal only under its local temporal context
  & Inject a local spike or drop within an otherwise smooth region so that it is inconsistent with nearby values. \\
\midrule
3 & Amplitude change & Seasonal
  & A change in the magnitude of a recurring or oscillatory pattern
  & Multiply a periodic segment by an amplitude scaling factor while preserving its frequency. \\
4 & Seasonality anomaly & Seasonal
  & A disruption or change in periodic structure or dominant frequency
  & Alter the period or frequency of a periodic segment, or partially break its regular repetition. \\
\midrule
5 & Trend change & Structural
  & A persistent change in the slope or long-term direction of the series
  & Add a piecewise linear drift or slope change after a selected point. \\
6 & Mean change point & Structural
  & An abrupt shift in the mean level of the series
  & Add a step-like level shift after a selected change point. \\
7 & Variance change & Structural
  & A persistent change in the variability or dispersion of the series
  & Increase or decrease the local noise variance within a segment. \\
\midrule
8 & Pattern shift & Pattern
  & A change in the recurring temporal shape or motif structure
  & Circularly shift the second half of the sequence by one quarter of the detected period. \\
9 & Waveform distortion & Pattern
  & A deformation of the original waveform while preserving rough temporal locality
  & Clip a local interval around its local mean to flatten waveform extrema. \\
\bottomrule
\end{tabularx}
\end{table}

\begin{table}[H]
\centering
\footnotesize
\setlength{\tabcolsep}{4pt}
\renewcommand{\arraystretch}{1.05}
\caption[Synthetic injection parameter settings for SAGE.]{Synthetic injection parameter settings used for the SAGE anomaly taxonomy. Here \(n\) denotes sequence length and \(\sigma\) denotes the standard deviation of the normal prototype.}
\label{tab:ch5_synthetic_injection_params}
\begin{tabularx}{0.96\textwidth}{@{}l>{\RaggedRight\arraybackslash}X@{}}
\toprule
Type & Parameterization \\
\midrule
Global point anomaly & Insert 1--3 isolated spikes or drops at random positions with magnitude \(5\sigma\). \\
Contextual point anomaly & Insert 1--3 local spikes using a local window of \(\max(10,n/20)\) and magnitude \(3\) times the local standard deviation. \\
Amplitude change & Scale the second-half deviation from the segment mean by a factor of \(2.0\). \\
Seasonality anomaly & Modify the second-half frequency using multiplier \(2.5\), or flatten the second-half seasonal amplitude to \(15\%\) of its original deviation and add Gaussian noise with standard deviation \(0.15\sigma\). \\
Trend change & Start at \(0.5n\) and add a linear trend with slope \(\max(0.05\sigma,0.05)\), reversing the direction when a strong pre-existing trend is detected. \\
Mean change point & Sample a change point uniformly within \([0.4n,0.6n]\) and shift the suffix by \(1.5\sigma\). \\
Variance change & Sample a change point within \([0.4n,0.6n]\) and scale suffix deviations from the segment mean by \(2.0\) or \(2.5\). \\
Pattern shift & Circularly shift the second half by one quarter of the detected period. \\
Waveform distortion & Clip the interval \([0.4n,0.7n]\) to the local mean \(\pm 0.5\) local standard deviations. \\
\bottomrule
\end{tabularx}
\end{table}

\begin{figure}[H]
\centering
\includegraphics[width=\linewidth]{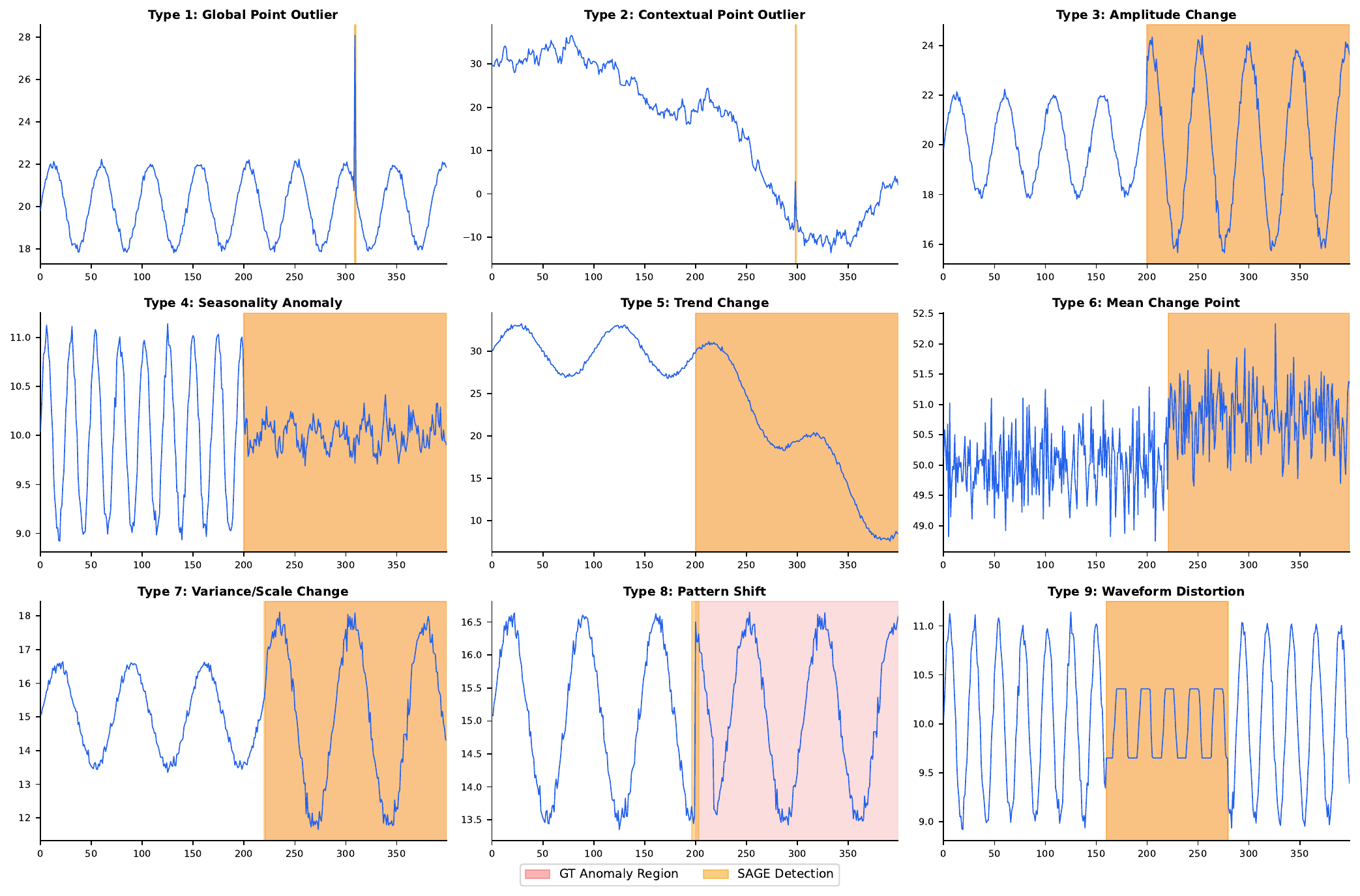}
\caption{Representative synthetic examples. Red shading marks the injected anomaly and orange shading the SAGE detection. The taxonomy covers isolated deviations, persistent changes, periodic disruptions, and shape distortions.}
\label{fig:synthetic_examples}
\end{figure}
\paragraph{Reference construction.}
Training sequences are partitioned into fixed-length segments. Segments overlapping binary-labeled anomalies are removed before prototype selection. The normal prototypes are observed subsequences selected with $k$-medoids; the number of medoids is chosen using a silhouette-based criterion, and small reference pools are retained without clustering. Each selected prototype produces one contrastive pair for each of the nine types. A pair stores the original and perturbed signals, the injected type, and a summary of statistical, change-point, spectral, autocorrelation, or segment-comparison evidence. DTW retrieval selects three normal prototypes, followed by type-based selection of their perturbations using Analyzer candidates. No test labels are used in these steps.

\paragraph{Type-evaluation aggregation.}
\label{app:type_aggregation}
Table~\ref{tab:types_family} preserves the reported family-level type-agreement summaries. Their aggregation cannot be reconstructed as a single pooled proportion from the displayed sample counts and recall rates. For example, 54\% recall among 24 seasonal samples implies 13 detected cases after rounding, whereas 75\% agreement cannot be obtained by rounding an integer number of correct types out of 13. The summaries do not establish whether finer-grained type rates were averaged. We therefore avoid inferring correct-type counts or uncertainty estimates from these percentages.

\section{Implementation and Experimental Settings}
\label{app:implementation}
The workflow uses LangGraph with parallel Analyzers, window size 400, stride 400, temperature 0.0, and three retrieved prototypes. Original floating-point values are retained for tools. The compact prompt contains global statistics, segment summaries, integer-rounded representative samples, and extrema-preserving indices. This separation preserves tool-level numerical precision while reducing the prompt representation of the signal.

Claude Sonnet 4.6 is used for the main framework comparison. Additional SAGE experiments use GPT-5.1, Claude Haiku 4.5, Qwen3.5-27B, and Qwen3.5-9B. Hosted backbones are accessed through APIs. Qwen3.5-27B uses vLLM with tensor parallelism across four NVIDIA A100 80GB GPUs, and Qwen3.5-9B uses one NVIDIA A100 80GB GPU. Runtime depends on the backbone, tool calls, and generated outputs; the workflow does not impose a common token or time budget across compared methods.

ML/DL baselines include Isolation Forest, OCSVM, LOF, LSTM-AE, LSTM-VAE, USAD, OmniAnomaly, and Anomaly Transformer \citep{liu2008isolation,scholkopf2001estimating,DBLP:conf/sigmod/BreunigKNS00,DBLP:journals/corr/MalhotraRAVAS16,DBLP:journals/ral/ParkHK18,DBLP:conf/kdd/AudibertMGMZ20,DBLP:conf/kdd/SuZNLSP19,DBLP:conf/iclr/XuWWL22}. LLM/VLM baseline reproduction retains the original representation, prompting, ICL, output format, and post-processing as closely as possible while using the common backbone.

\subsection{Compact input and synthetic reference selection}
\label{app:compact_input}
The detailed configuration recorded for the original implementation uses a 400-point window and an agent-dependent compact prompt budget of 200--500 estimated tokens. Each window retains its length, range, mean, standard deviation, and segment-level statistics. If the numerical sequence exceeds the prompt budget, representative indices are sampled at a uniform stride determined by the estimated budget; the global minimum and maximum indices are appended if absent. Sampled values are rounded to integers in the prompt, while all numerical tools operate on the original floating-point signal.

For reference construction, let $N$ be the number of eligible normal-reference segments after excluding windows overlapping labeled training anomalies. When $N\leq12$, all segments are retained. Otherwise, the number of $k$-medoids prototypes is selected by the silhouette score over $k\in[2,\min(20,\lfloor N/2\rfloor)]$. One perturbation per anomaly type is generated for each selected prototype. At inference, DTW with LB\_Keogh pruning selects three normal prototypes; Analyzer-predicted types then select the associated contrastive pairs. The prototype count $k$ and the retrieval count of three are separate settings.

\subsection{Numerical tool configuration}
\label{app:tool_parameters}
Table~\ref{tab:tool_parameters} lists the documented numerical configuration. Parameters are fixed across datasets unless inferred from the window length or a detected period. These are tool settings, distinct from the Detector's evidence-strength rubric. The values are retained from the original implementation record; the updated score summary does not independently document a new parameter configuration.

\begin{table}[H]
\centering
\footnotesize
\caption{Recorded numerical tool and reference settings.}
\label{tab:tool_parameters}
\renewcommand{\arraystretch}{1.12}
\begin{tabularx}{\linewidth}{@{}p{0.31\linewidth}X@{}}
\toprule
Component & Configuration \\
\midrule
Global point candidates & Z-score threshold 3.0; IQR multiplier 1.5. \\
Contextual point candidates & Multi-scale rolling statistics; local Z-score threshold 2.5. \\
Decomposition & FFT-based period estimation when unspecified; moving-average or STL-style decomposition according to the tool call. \\
Symbolic representation & SAX alphabet size 10. \\
Transition representation & MTF bin count 10. \\
Recurrence plot & Neighborhood percentile parameter 0.1. \\
Time--frequency analysis & STFT window 64; hop 32. \\
Regime expansion & Verification chunks of 50 points; reference window of 100 points; mean/variance difference tests at $p<0.05$. \\
Prototype selection & Keep all segments for $N\leq12$; otherwise select $k$ by silhouette over $[2,\min(20,\lfloor N/2\rfloor)]$. \\
Synthetic retrieval & Three normal prototypes by DTW with LB\_Keogh pruning, followed by Analyzer-type-based contrast selection. \\
\bottomrule
\end{tabularx}
\end{table}

\subsection{Software and inference environment}
\label{app:software}
The recorded software environment is Python 3.10.12, PyTorch 2.10.0 with CUDA 12.8, vLLM 0.17.1, LangChain 0.3.27, LangGraph 0.6.7, Anthropic SDK 0.77.0, and OpenAI SDK 2.16.0. Sonnet 4.6, Haiku 4.5, and GPT-5.1 are accessed through hosted APIs. Qwen3.5-27B is served in FP16 with tensor parallel size 4 on four NVIDIA A100 80GB GPUs; Qwen3.5-9B is served in BF16 on one NVIDIA A100 80GB GPU. These versions and precision settings describe the environment associated with the supplementary runtime measurements; equivalence to the main evaluation configuration is not established by the available records.

\subsection{Recorded inference time and token usage}
\label{app:runtime}
\begin{table}[!htbp]
\centering
\footnotesize
\caption{Recorded Yahoo S5 inference time per 400-point window. Values are approximate means from the original experiment logs; execution setups are not matched across backbones.}
\label{tab:efficiency}
\begin{tabular}{@{}llr@{}}
\toprule
Backbone & Execution setup & Seconds/window \\
\midrule
Claude Sonnet 4.6 & Hosted API & 80 \\
GPT-5.1 & Hosted API & 34 \\
Qwen3.5-27B & $4\times$ A100 80GB, FP16 & 481 \\
Qwen3.5-9B & $1\times$ A100 80GB, BF16 & 44 \\
\bottomrule
\end{tabular}
\end{table}

Table~\ref{tab:efficiency} reports supplementary Yahoo S5 runtime measurements. The execution environments and their relation to the main experiments are described in Appendix~\ref{app:software}. The hosted Sonnet and GPT runs averaged approximately 80 and 34 seconds per 400-point window, respectively. Qwen3.5-27B averaged 481 seconds on four A100 GPUs, while Qwen3.5-9B averaged 44 seconds on one A100. Because API serving and local inference use different hardware and scheduling, these values are observed workflow latencies rather than a controlled comparison of model efficiency. The retained summary does not report runtime dispersion, a comparable Haiku runtime, or a common serving budget; no deployment throughput guarantee or monetary cost ranking follows from these numbers.

Processing throughput and alert-response latency are distinct. Window arrival rates depend on sampling cadence, stride, and the number of monitored streams; alerts may also wait for a window to fill before inference starts. An 80-second inference delay is unsuitable for second-level response requirements, even if average processing throughput keeps pace. The current implementation is therefore most relevant when detailed diagnosis can tolerate delayed responses.

A representative Sonnet 4.6 run on Yahoo A1 used approximately 22.4K input tokens and 4.9K output tokens per window. This Yahoo A1 observation is not an average across datasets or backbones. It is a workflow-level measurement, distinct from the 200--500 estimated-token compact signal summaries supplied to individual agent roles (Appendix~\ref{app:implementation}). Compressing the numerical input therefore does not remove the cost of multiple calls and intermediate evidence. The measurements motivate a two-stage deployment in which a lightweight screen selects suspicious windows for detailed diagnosis; the accuracy and cost of such screening remain to be evaluated, including anomalies missed by the first-stage filter.

\Needspace{0.48\textheight}
\section{Analyzer Tool Assignments}
\label{app:tools}
The primary tools generate family-specific evidence, while shared auxiliary tools allow complementary checks. The agent roles and their permitted types are specified in Appendix~\ref{app:prompts}.

\begin{table}[H]
\centering
\footnotesize
\setlength{\tabcolsep}{2.4pt}
\renewcommand{\arraystretch}{0.88}
\caption{Main tools used by the four Analyzers. P denotes a primary tool for the corresponding Analyzer, and S denotes a shared auxiliary tool.}
\label{tab:ch5_analyzer_tools}
\begin{tabularx}{\textwidth}{@{}>{\RaggedRight\arraybackslash}p{0.13\textwidth}>{\RaggedRight\arraybackslash}p{0.27\textwidth}>{\RaggedRight\arraybackslash}Xc@{}}
\toprule
Analyzer & Tool & Function & Role \\
\midrule
\multirow{5}{*}{Point}
& \texttt{statistics} & Mean, standard deviation, skewness, kurtosis & P \\
& \texttt{outliers} & Global Z-score and IQR-based outlier detection & P \\
& \texttt{rolling\_statistics} & Multi-scale rolling statistics & P \\
& \texttt{symbolic\_repr} & SAX-based symbolic representation & S \\
& \texttt{recurrence\_plot} & Recurrence-plot-based structural analysis & S \\
\midrule
\multirow{6}{*}{Structural}
& \texttt{decomposition} & STL-based trend/seasonal/residual decomposition & P \\
& \texttt{diff} & First- and second-order differences & P \\
& \texttt{change\_point} & CUSUM-based change-point detection & P \\
& \texttt{segment\_comparison} & Segment-level statistical comparison & P \\
& \texttt{GAF} & Gramian Angular Field visual representation & S \\
& \texttt{recurrence\_plot} & Recurrence-plot-based structural analysis & S \\
\midrule
\multirow{7}{*}{Seasonal}
& \texttt{auto\_correlation} & Split-half autocorrelation comparison & P \\
& \texttt{fourier\_transform} & FFT-based frequency component extraction & P \\
& \texttt{wavelet\_transform} & Wavelet-based time-frequency analysis & P \\
& \texttt{stft\_analysis} & Short-time Fourier transform analysis & P \\
& \texttt{symbolic\_repr} & SAX-based symbolic representation & S \\
& \texttt{recurrence\_plot} & Recurrence-plot-based structural analysis & S \\
& \texttt{MTF} & Markov Transition Field visual representation & S \\
\midrule
\multirow{6}{*}{Pattern}
& \texttt{symbolic\_repr} & SAX-based symbolic representation & P \\
& \texttt{recurrence\_plot} & Recurrence-pattern analysis & P \\
& \texttt{rolling\_statistics} & Local pattern statistics & S \\
& \texttt{auto\_correlation} & Split-half autocorrelation comparison & S \\
& \texttt{GAF} & Gramian Angular Field visual representation & P \\
& \texttt{MTF} & Markov Transition Field visual representation & P \\
\bottomrule
\end{tabularx}
\end{table}

\section{Agent Prompt Specifications}
\label{app:prompts}
The following prompts specify the experimental agent roles. They describe intended behavior rather than enforceable guarantees. In particular, the Detector prompt permits additional chart-supported candidates, whereas the Supervisor is instructed to preserve confirmed records.

\begin{promptbox}[Global Rules (shared)]
\begin{enumerate}[nosep,leftmargin=*]
\item A data point is considered anomalous if it is part of a consecutive anomaly sequence, or if it shows a sharp drop or spike.
\item A data point is considered anomalous if it remains below/above predefined normal thresholds for an extended period and falls outside the expected normal range.
\item Typically less than 5\% of points are anomalous. An empty anomaly list is expected and valid in most windows.
\item Normal data can exhibit variability, which must not be mistaken for anomalies.
\item Misclassifying normal data as anomalous can lead to critical failures. Exercise extreme caution. False positives are not tolerated.
\item Only flag data as anomalous when supported by strong statistical or visual evidence.
\item Anomaly interval outputs must be precisely located and must not be excessively long.
\item Merge nearby anomalous points into segments. Output only segments with strong evidence.
\item An empty anomaly list [] is valid and common. Do NOT fabricate anomalies to meet a quota.
\item When no analyzer candidate has strong supporting evidence, return an empty list.
\end{enumerate}
\end{promptbox}

\begin{agentbox}[PointAnalyzer --- Types 1, 2]
You are a point-level anomaly detection specialist. Your task is to identify potential point anomalies in time series data.

\textbf{Your Scope (Types 1, 2 ONLY)}\\
- Type 1: Point Outlier (Global) -- Values exceeding global Z-score threshold ($|Z| \geq 3$), including clustered noise\\
- Type 2: Point Outlier (Contextual) -- Values abnormal relative to local context

\textbf{Visual Analysis}\\
When a chart image is provided, visually confirm whether flagged points truly stand out from surrounding values. Use the image to verify spike magnitude and local context.

\textbf{Analysis Steps}\\
1. Compute global statistics (mean, std, IQR) for Z-score thresholds\\
2. Run statistical outlier detection to find candidate indices\\
3. Compute rolling statistics to identify contextual anomalies\\
4. Classify each candidate as Type 1 or 2

\textbf{Output}: List candidate outlier indices with scores, statistics summary, candidate anomaly types from \{1, 2\}. Do NOT attempt structural or seasonal analysis.
\end{agentbox}

\begin{agentbox}[SeasonAnalyzer --- Types 3, 4]
You are a seasonal and frequency anomaly detection specialist. Your task is to identify seasonal/frequency anomalies in time series data.

\textbf{Your Scope (Types 3, 4 ONLY)}\\
- Type 3: Amplitude Change -- Change in seasonal amplitude\\
- Type 4: Seasonality Anomaly -- Distortion of seasonal pattern, frequency/period change

\textbf{Visual Analysis}\\
When a chart image is provided, visually confirm seasonal pattern changes. Look for visible amplitude reduction/increase, period changes, or loss of regular oscillation. Use the image to identify the full extent of the affected region.

\textbf{Analysis Steps}\\
1. Run autocorrelation analysis to detect dominant periods\\
2. Apply Fourier transform for frequency domain analysis\\
3. Apply wavelet transform for time-frequency localization\\
4. Compare expected vs observed seasonal patterns

\textbf{Output}: List dominant periods detected, frequency-domain anomalies, seasonality summary, candidate anomaly types from \{3, 4\}. Do NOT analyze point outliers or structural changes.
\end{agentbox}

\begin{agentbox}[StructAnalyzer --- Types 5, 6, 7]
You are a structural change detection specialist. Your task is to identify structural changes in time series data.

\textbf{Your Scope (Types 5, 6, 7 ONLY)}\\
- Type 5: Trend Change -- Trend direction change\\
- Type 6: Change Point (Mean) -- Statistical mean change point, persistent level shift\\
- Type 7: Variance/Scale Change -- Variance or overall scale change point

\textbf{Visual Analysis}\\
When a chart image is provided, visually confirm structural changes. Look for visible level shifts, trend reversals, or variance changes. Use the image to determine the full extent of affected regions after change points.

\textbf{Analysis Steps}\\
1. Decompose time series (trend, seasonality, residual)\\
2. Compute first/second order differences for trend analysis\\
3. Run change point detection (statistical tests)\\
4. Compare segments before/after detected change points

\textbf{Output}: List detected change point indices, trend behavior description, segment comparison results, candidate anomaly types from \{5, 6, 7\}. Do NOT analyze point outliers or seasonal patterns.
\end{agentbox}

\begin{agentbox}[PatternAnalyzer --- Types 8, 9]
You are a pattern-based anomaly detection specialist. Your task is to identify pattern-level anomalies in time series data using symbolic representations and recurrence analysis.

\textbf{Your Scope (Types 8, 9 ONLY)}\\
- Type 8: Pattern Shift -- Change in the shape/phase of repeating patterns\\
- Type 9: Waveform Distortion -- Distortion in repeating waveform shape

\textbf{Visual Analysis}\\
When chart images (time series, GAF, MTF, recurrence plot) are provided, visually confirm pattern changes. Look for visible breaks in repeating structure, waveform shape changes, or regime transitions.

\textbf{Analysis Steps}\\
1. Analyze SAX symbolic representation for pattern repetitions and breaks\\
2. Examine recurrence plot metrics for regime transitions and determinism changes\\
3. Use autocorrelation to confirm pattern repetition periods\\
4. Use rolling statistics to detect local pattern changes (variance shifts)\\
5. Synthesize all evidence to identify pattern anomalies

\textbf{Key Indicators}: SAX pattern breaks (sudden symbolic sequence changes), recurrence plot low determinism/laminarity (loss of pattern structure), autocorrelation period changes, local variance spikes.

\textbf{Output}: List symbolic pattern repetitions and break points, recurrence analysis metrics, candidate pattern anomaly indices, candidate anomaly types from \{8, 9\}. Do NOT analyze point outliers, structural changes, or seasonal patterns.
\end{agentbox}

\begin{detectorbox}[Detector --- Evidence Aggregation]
You are an expert in time series anomaly detection. Your task is to score candidate anomalous regions with an integer confidence value (0 to 100).

\textbf{Scoring Rubric (0--100 integer)}\\
Score based on EVIDENCE STRENGTH from any source: tool results, raw data inspection, or visual chart analysis.

\begin{tabularx}{\linewidth}{@{}lX@{}}
0--10 & Clearly normal variation. No anomaly from any perspective.\\
10--30 & Weak evidence (minor deviation, ambiguous visual pattern).\\
30--50 & Borderline (small isolated deviation that could be noise).\\
50--70 & Clear anomaly: obvious spike/dip OR z${>}$2 OR structural change.\\
70--85 & Strong: large magnitude ($\geq$2$\times$ background) OR multi-point cluster.\\
85--100 & Overwhelming: extreme magnitude OR confirmed by multiple sources.\\
\end{tabularx}

\textbf{Scoring Rules}\\
- Score reflects evidence clarity, NOT just z-score number.\\
- Magnitude over z-score: check actual value vs background range.\\
- Spike cluster bonus: 2+ consecutive $\to$ $\geq$60; 3+ $\to$ $\geq$70.\\
- Do NOT output anomalies with score below 50.\\
- Index precision is critical: use tool-reported indices directly.\\
- Beyond analyzer candidates: if the data or chart reveals missed anomalies, report them.

\textbf{Visual Analysis} (when image provided)\\
Examine the time series chart to visually confirm anomalies. For regime changes, mark the ENTIRE affected region after the change point.

\textbf{Output}: List of \{index, end\_index, confidence, types\}. Empty list if no anomalies.
\end{detectorbox}

\begin{supervisorbox}[Supervisor --- Final Diagnosis Report]
You are the orchestrator of a time-series anomaly detection system. Your task is to convert structured anomaly records and supporting evidence into an analyst-facing diagnosis report.

\textbf{Role}\\
- Coordinate the workflow at the start of analysis.\\
- Generate the final report when anomaly records are confirmed.\\
- Generate a ``no anomaly'' report when no anomaly record is confirmed.\\
- Do NOT introduce new anomaly intervals, anomaly types, or unsupported causes beyond the Detector output and Analyzer evidence.

\textbf{Report Generation Guidelines}\\
\textbf{Executive Summary}: Summarize the key result in 1--2 sentences, including whether anomalies were found, their severity, and whether immediate action is required.\\
\textbf{Time-Series Characteristics}: Describe overall trend, seasonality, data quality, notable features, normal range, and variability.\\
\textbf{Anomaly Details}: For each confirmed anomaly, report its location, type, description, confidence, severity, and impact scope.\\
\textbf{Alarm Level}: Determine the overall alarm level according to the predefined alarm-level rubric.\\
\textbf{Recommendations}: Provide immediate actions for Urgent/Error cases, monitoring strategies, and follow-up analysis suggestions.

\textbf{Output Format}\\
The final report must include the following fields:\\
- \texttt{executive\_summary}: Executive summary.\\
- \texttt{time\_series\_characteristics}: Time-series characteristics.\\
- \texttt{confirmed\_anomalies}: List of confirmed anomalies.\\
- \texttt{overall\_alarm\_level}: Overall alarm level.\\
- \texttt{alarm\_reason}: Reason for alarm-level determination.\\
- \texttt{recommendations}: List of recommendations.
\end{supervisorbox}

\section{Additional Detection Results}
\label{app:additional_results}
The complete Yahoo S5, KPI, and WSD benchmark results and their averages appear together in Table~\ref{tab:main}. This appendix reports analyses not contained in that table.

\subsection{Supplementary lightweight-detector comparison}
\label{app:lightweight}
The supplementary experimental record compares Isolation Forest, Matrix Profile implemented with STUMPY, and STL residual analysis. It reports independent min--max normalization of the three score sequences within each time series, followed by equal-weight average-score fusion. Table~\ref{tab:lightweight} preserves that record as a complete comparison, separately from the updated benchmark tables.

\begin{table}[H]
\centering
\small
\caption{Supplementary lightweight-detector comparison from the earlier experimental record: Point-F1 (percent). Values preserve the original reporting precision; the SAGE row belongs to this comparison rather than replacing the updated results in Table~\ref{tab:main}.}
\label{tab:lightweight}
\begin{tabular}{@{}lrrrr@{}}
\toprule
Method & Yahoo & KPI & WSD & Average \\
\midrule
Isolation Forest & 51.4 & 39.9 & 33.1 & 41.5 \\
Matrix Profile & 11.9 & 7.7 & 11.0 & 10.2 \\
STL Residual & 72.0 & 35.3 & 52.2 & 53.2 \\
Average-score ensemble & 64.6 & 27.5 & 37.2 & 43.1 \\
SAGE & \textbf{79.7} & \textbf{61.0} & \textbf{58.1} & \textbf{66.3} \\
\bottomrule
\end{tabular}
\end{table}

SAGE exceeds the strongest individual lightweight detector, STL Residual, by 13.1 percentage points on average in this record, and the average-score ensemble by 23.2 points. The ensemble falls below STL alone on all three datasets. Thus, naive averaging does not reliably combine these heterogeneous scores; the result does not establish superiority over learned, calibrated, or otherwise improved lightweight fusion. Nor does it separate the benefit of the LLM from SAGE's different tools, visual inputs, and reference construction. The human study compares SAGE with LLM/VLM systems, not with these lightweight alternatives.

The retained comparison summary does not specify the Matrix Profile subsequence length, STL period, or the time range used to fit the score normalization. It also lacks a recoverable numerical CPU runtime and a complete hardware description. We therefore do not report an efficiency ratio or treat its normalization as a verified online procedure. These missing settings limit reproducibility of this supplementary comparison and must be resolved before drawing deployment conclusions.

\Needspace{0.30\textheight}
\subsection{Yahoo subset results}
Table~\ref{tab:subset_yahoo} compares the real A1 subset and synthetic A2--A4 subsets. SAGE has the strongest Point-F1 on A1 and A4; LLMAD leads on A2 and A3.

\begin{table}[H]
\centering
\footnotesize
\caption{Yahoo S5 subset results (percent). A1 contains real traffic; A2--A4 are synthetic. Pt, PA, Aff, and Del denote Point-F1, PA-F1, Affiliation-F1, and Delayed-F1; column maxima are bold.}
\label{tab:subset_yahoo}
\resizebox{\textwidth}{!}{%
\begin{tabular}{@{}l|cccc|cccc|cccc|cccc@{}}
\toprule
\multirow{2}{*}{Method}
& \multicolumn{4}{c}{A1} & \multicolumn{4}{c}{A2}
& \multicolumn{4}{c}{A3} & \multicolumn{4}{c}{A4} \\
\cmidrule(lr){2-5}\cmidrule(lr){6-9}\cmidrule(lr){10-13}\cmidrule(lr){14-17}
& Pt & PA & Aff & Del & Pt & PA & Aff & Del
& Pt & PA & Aff & Del & Pt & PA & Aff & Del \\
\midrule
SigLLM & 36.19 & 42.62 & 81.52 & 33.83 & 48.01 & 48.01 & 86.00 & 44.01 & 73.33 & 73.39 & \textbf{92.58} & 80.43 & 50.94 & 50.94 & 83.59 & 64.24 \\
LLMAD & 54.03 & 55.82 & 85.93 & 74.68 & \textbf{89.03} & \textbf{89.03} & \textbf{93.83} & \textbf{91.47} & \textbf{85.72} & \textbf{85.72} & 92.43 & \textbf{88.63} & 68.93 & 68.93 & 82.23 & \textbf{74.83} \\
TAMA & 10.44 & 10.97 & 66.38 & 65.80 & 2.58 & 2.74 & 72.36 & 56.10 & 3.22 & 3.22 & 64.79 & 58.22 & 1.77 & 1.77 & 65.41 & 66.29 \\
TSAD-Agents & 26.87 & 28.37 & 60.92 & 33.33 & 41.98 & 41.98 & 59.68 & 42.77 & 31.44 & 31.44 & 65.87 & 42.11 & 25.44 & 25.44 & 59.78 & 35.49 \\
SAGE & \textbf{75.31} & \textbf{78.42} & \textbf{88.37} & \textbf{90.46} & 86.48 & 87.13 & 89.02 & 88.32 & 83.27 & 83.27 & 90.51 & 85.63 & \textbf{70.72} & \textbf{70.72} & \textbf{85.61} & 74.51 \\
\bottomrule
\end{tabular}
}
\end{table}

The A1 Point-F1 gap between SAGE and LLMAD is 21.28 percentage points, whereas the two methods are much closer on the synthetic A2--A4 subsets. The four metrics do not always rank methods identically: on A4, SAGE leads LLMAD in Point-F1 and Affiliation-F1, while LLMAD has a slightly higher Delayed-F1. The pooled Yahoo result in Table~\ref{tab:main} therefore should not be taken as uniform performance across its real and synthetic subsets.

\Needspace{0.60\textheight}
\subsection{Ablations and zero-shot comparisons}
These experiments remove one component at a time. On KPI and WSD, removing specialization improves some relaxed segment metrics even though the full system retains the strongest Point-F1.
The available experimental description does not establish a matched token budget or fully specify the remaining agent structure and tool-access policy in the single-analyzer condition. We therefore interpret the ablations as comparisons of the recorded workflow variants, not an isolated causal estimate of the benefit of adding agents at fixed compute.

\begin{table}[H]
\centering
\footnotesize
\setlength{\tabcolsep}{3pt}
\renewcommand{\arraystretch}{0.95}
\caption{Detailed ablation results. Each row removes one component from the full SAGE system. Values are percentages, and best results are shown in \textbf{bold}.}
\label{tab:ch5_ablation_all}
\begin{tabular}{@{}llcccc@{}}
\toprule
\multirow{2}{*}{Benchmark} & \multirow{2}{*}{Setting} & \multicolumn{4}{c}{Metrics} \\
\cmidrule(lr){3-6}
& & Pt & PA & Aff & Del \\
\midrule
\multirow{5}{*}{Yahoo Overall}
& Full             & \textbf{79.69} & \textbf{80.46} & \textbf{88.37} & \textbf{84.32} \\
& - Multi-analyzer & 41.08 & 42.93 & 81.57 & 58.23 \\
& - Tools          & 34.79 & 36.72 & 74.68 & 54.53 \\
& - ICL            & 71.21 & 72.03 & 87.48 & 81.32 \\
& - Vision         & 67.33 & 68.12 & 83.78 & 77.03 \\
\midrule
\multirow{5}{*}{KPI}
& Full             & \textbf{60.97} & 78.61 & 86.42 & 74.83 \\
& - Multi-analyzer & 59.22 & \textbf{79.58} & \textbf{91.12} & \textbf{76.83} \\
& - Tools          & 44.38 & 68.47 & 81.63 & 70.58 \\
& - ICL            & 49.21 & 64.87 & 78.53 & 67.58 \\
& - Vision         & 47.83 & 61.72 & 79.87 & 69.13 \\
\midrule
\multirow{5}{*}{WSD}
& Full             & \textbf{58.12} & 95.68 & 96.82 & 97.08 \\
& - Multi-analyzer & 55.54 & \textbf{98.87} & \textbf{97.83} & \textbf{98.07} \\
& - Tools          & 56.07 & 96.17 & 97.28 & 97.03 \\
& - ICL            & 54.87 & 94.43 & 95.57 & 95.23 \\
& - Vision         & 52.97 & 92.73 & 95.87 & 95.93 \\
\bottomrule
\end{tabular}
\end{table}

Table~\ref{tab:ch5_ablation_all} shows that removing tools or specialization on Yahoo reduces both Point-F1 and PA-F1 by more than 37 percentage points, while the corresponding Affiliation-F1 declines are smaller. On KPI and WSD, removing specialization slightly lowers Point-F1 but raises several segment-oriented metrics. The variants therefore affect precise localization and broader interval coverage differently. Each row is a separate removal from the full workflow; differences between rows do not measure interactions among components.

\begin{table}[H]
\centering
\small
\caption{Comparison of ICL and zero-shot settings for SAGE and LLMAD. Values are percentages.}
\label{tab:ch5_zeroshot_llmad}
\begin{tabular}{llcccc}
\toprule
Dataset & Method & Pt & PA & Aff & Del \\
\midrule
\multirow{4}{*}{Yahoo}
& LLMAD (ICL)       & 76.82 & 77.12 & 88.93 & 83.38 \\
& LLMAD (Zero-shot) & 76.28 & 76.83 & \textbf{90.27} & 81.73 \\
& SAGE (ICL)        & \textbf{79.69} & \textbf{80.46} & 88.37 & \textbf{84.32} \\
& SAGE (Zero-shot)  & 71.21 & 72.03 & 87.48 & 81.32 \\
\midrule
\multirow{4}{*}{KPI}
& LLMAD (ICL)       & 19.47 & 44.21 & 84.66 & 48.02 \\
& LLMAD (Zero-shot) & 10.87 & 28.52 & 77.83 & 44.07 \\
& SAGE (ICL)        & \textbf{60.97} & \textbf{78.61} & \textbf{86.42} & \textbf{74.83} \\
& SAGE (Zero-shot)  & 49.21 & 64.87 & 78.53 & 67.58 \\
\midrule
\multirow{4}{*}{WSD}
& LLMAD (ICL)       & 41.59 & 71.54 & 79.36 & 77.04 \\
& LLMAD (Zero-shot) & 24.43 & 41.57 & 78.03 & 67.42 \\
& SAGE (ICL)        & \textbf{58.12} & \textbf{95.68} & \textbf{96.82} & \textbf{97.08} \\
& SAGE (Zero-shot)  & 54.87 & 94.43 & 95.57 & 95.23 \\
\bottomrule
\end{tabular}
\end{table}

Table~\ref{tab:ch5_zeroshot_llmad} separates the contribution of retrieved references from the full-system comparison. Removing ICL lowers SAGE Point-F1 by 8.48, 11.76, and 3.25 points on Yahoo, KPI, and WSD, respectively. The effect is not uniform across methods or metrics: LLMAD's Yahoo Affiliation-F1 is slightly higher without ICL, despite a small Point-F1 decrease. Zero-shot SAGE still exceeds LLMAD with ICL in Point-F1 on KPI and WSD, but not on Yahoo. These comparisons do not isolate which feature of the synthetic references produces the gains.

\Needspace{0.38\textheight}
\subsection{Backbone results}
Table~\ref{tab:ch5_backbone_details} reports the updated Point-F1 comparison. The full earlier backbone record is preserved separately below.

\begin{table}[H]
\centering
\footnotesize
\setlength{\tabcolsep}{2pt}
\renewcommand{\arraystretch}{1.0}
\caption{Backbone-wise Point-F1 results of SAGE across three benchmarks. Values are percentages, and best results are shown in \textbf{bold}.}
\label{tab:ch5_backbone_details}
\begin{tabular}{@{}lrrrr@{}}
\toprule
Model & Yahoo & KPI & WSD & Avg. Point-F1 \\
\midrule
\textbf{Sonnet 4.6} & \textbf{79.69} & \textbf{60.97} & \textbf{58.12} & \textbf{66.26} \\
GPT-5.1 & 50.87 & 39.13 & 54.61 & 48.20 \\
Qwen 3.5-27B & 54.52 & 27.48 & 46.53 & 42.84 \\
Haiku 4.5 & 42.91 & 22.28 & 46.47 & 37.22 \\
Qwen 3.5-9B & 7.63 & 3.88 & 20.19 & 10.57 \\
\bottomrule
\end{tabular}
\end{table}

Sonnet 4.6 leads the other evaluated backbones in Point-F1 on all three benchmarks. Its average of 66.26 contrasts with 48.20 for the next highest average, GPT-5.1; the gap is larger on KPI than on WSD. The range from 10.57 to 66.26 average Point-F1 shows that the reported detection performance depends substantially on the backbone used within the same workflow. This comparison does not by itself assess whether the resulting diagnostic explanations remain equally useful across backbones.

\Needspace{0.48\textheight}
\subsubsection{Supplementary full-metric backbone record}
\label{app:backbone_full}
Table~\ref{tab:backbone_full} reports all four metrics from the earlier backbone evaluation. These values were originally reported to three decimal places on a 0--1 scale and are shown here to one decimal place as percentages. The updated Point-F1 comparison remains Table~\ref{tab:ch5_backbone_details}, and the current Sonnet results across all four metrics remain in Table~\ref{tab:main}. In particular, the earlier Sonnet values for some segment metrics differ from the updated results; these earlier measurements are therefore reported separately.

\begin{table}[H]
\centering
\footnotesize
\caption{Supplementary full-metric backbone results from the earlier evaluation record (percent, preserving the original reporting precision). This record is kept intact rather than merged with the updated Point-F1 summary in Table~\ref{tab:ch5_backbone_details}.}
\label{tab:backbone_full}
\begin{tabular}{@{}llrrrr@{}}
\toprule
Dataset & Backbone & Point-F1 & PA-F1 & Affiliation-F1 & Delayed-F1 \\
\midrule
\multirow{5}{*}{Yahoo}
& Sonnet 4.6 & 79.7 & 80.4 & 88.4 & 84.3 \\
& GPT-5.1 & 50.9 & 52.2 & 69.2 & 61.2 \\
& Qwen3.5-27B & 54.5 & 55.5 & 70.9 & 64.7 \\
& Haiku 4.5 & 42.9 & 43.6 & 73.5 & 70.0 \\
& Qwen3.5-9B & 7.6 & 7.7 & 76.2 & 49.5 \\
\midrule
\multirow{5}{*}{KPI}
& Sonnet 4.6 & 61.0 & 79.1 & 87.1 & 73.4 \\
& GPT-5.1 & 39.1 & 54.8 & 77.1 & 56.8 \\
& Qwen3.5-27B & 27.5 & 41.8 & 75.8 & 54.1 \\
& Haiku 4.5 & 22.3 & 35.7 & 71.6 & 55.4 \\
& Qwen3.5-9B & 3.9 & 8.5 & 79.4 & 17.3 \\
\midrule
\multirow{5}{*}{WSD}
& Sonnet 4.6 & 58.1 & 95.7 & 97.4 & 97.0 \\
& GPT-5.1 & 54.6 & 94.0 & 97.2 & 97.0 \\
& Qwen3.5-27B & 46.5 & 83.5 & 87.3 & 90.4 \\
& Haiku 4.5 & 46.5 & 76.1 & 79.6 & 88.9 \\
& Qwen3.5-9B & 20.2 & 26.5 & 87.5 & 50.8 \\
\bottomrule
\end{tabular}
\end{table}

The additional metrics show a distinction between precise localization and approximate temporal proximity. For example, Qwen3.5-9B has WSD Point-F1 of 20.2 but Affiliation-F1 of 87.5 in the earlier record; on Yahoo, the corresponding scores are 7.6 and 76.2. GPT-5.1 attains WSD PA-F1 of 94.0 and Delayed-F1 of 97.0, while its Yahoo and KPI scores are lower. High proximity or overlap scores alongside low Point-F1 should not be interpreted as precise anomaly boundaries. These observations describe the earlier recorded metric profiles; they do not establish that a smaller backbone provides equivalent diagnostic reports.

\Needspace{0.48\textheight}
\section{Auxiliary Benchmark Type Annotation}
\label{app:annotation}
One annotator assigned primary and secondary anomaly types to 626 intervals in 100 anomalous files selected across datasets using visual inspection and statistical analysis. The reported results use the 300 intervals in the test split and compare the primary annotation with SAGE's type prediction. Of these, 153 are global point anomalies, 98 contextual point anomalies, 39 mean change points, and 10 pattern shifts. This single-annotator, imbalanced subset is a qualitative supplement; it is not a balanced ground-truth benchmark for all nine types. The aggregate and per-type agreement rates in Table~\ref{tab:ch5_type_agreement_results} are preserved as reported, but do not reconcile as one pooled count: the rounded per-type rates imply 295 correct labels out of 300 (98.3\%), whereas the reported overall rate is 98.0\%. Without interval-level predictions and an explicit aggregation rule, we do not use these summaries as a precise estimate of diagnostic accuracy.

\begin{table}[H]
\centering
\setlength{\tabcolsep}{4pt}
\renewcommand{\arraystretch}{0.95}
\caption{Author-annotated benchmark type distribution and reported agreement summaries. Aggregate and per-type rates are not fully reconciled; see the text for the aggregation limitation.}
\label{tab:ch5_type_agreement_results}
\begin{subtable}[t]{0.46\textwidth}
\centering
\footnotesize
\caption{Overall type agreement}
\label{tab:ch5_type_agreement_overall}
\begin{tabular}{@{}lcccc@{}}
\toprule
Model & Overall & Yahoo & KPI & WSD \\
\midrule
Sonnet 4.6 & \textbf{98.0\%} & 97.7\% & 98.5\% & 100.0\% \\
GPT-5.1 & 84.7\% & 80.8\% & 91.8\% & 100.0\% \\
\bottomrule
\end{tabular}
\end{subtable}
\hfill
\begin{subtable}[t]{0.46\textwidth}
\centering
\footnotesize
\caption{Per-type agreement}
\label{tab:ch5_type_agreement_per_type}
\begin{tabular}{@{}lrc@{}}
\toprule
Anomaly Type & N & Agreement \\
\midrule
Global point anomaly & 153 & 99.3\% \\
Contextual point anomaly & 98 & 95.9\% \\
Mean change point & 39 & 100.0\% \\
Pattern shift & 10 & 100.0\% \\
\midrule
Overall & 300 & 98.0\% \\
\bottomrule
\end{tabular}
\end{subtable}
\end{table}

Point anomalies account for 251 of the 300 evaluated intervals, so the overall agreement is driven largely by the two point categories. The 10 pattern-shift intervals provide much less evidence about that type, and five of the nine taxonomy types are absent from this subset. The 100\% WSD entries in the model-level table likewise should be read in light of the unspecified WSD sample count. The controlled synthetic evaluation in Table~\ref{tab:types} covers all nine types and provides the broader type-level assessment.

\section{Human Evaluation Protocol and Detailed Results}
\label{app:human}
Ten technically experienced evaluators rated 25 samples per method: eight Yahoo S5 examples sampled across A1--A4, four randomly selected anomalous KPI files, four randomly selected anomalous WSD files, and nine synthetic examples (one per type). They saw the input series, true and predicted intervals, and anonymized diagnostic explanations in randomized order. Predictions were unchanged from the quantitative evaluation, so ratings are conditional on those outputs. Table~\ref{tab:ch5_human_eval_rubric} defines the four 1--5 scoring criteria, and Figure~\ref{fig:human_interface} shows the interface.

\subsection{Evaluator-level scores}
Table~\ref{tab:raters_all} reports each evaluator's mean score over the 25 selected cases for each method and criterion. SAGE has the highest overall average for nine evaluators and ties LLMAD for the tenth (E4). Its evaluator-level overall means range from 3.64 to 4.73, showing that individual assessments still vary. This pattern complements the aggregate ratings but remains descriptive for the selected cases.
\begin{landscape}
\begin{table}[H]
\centering
\scriptsize
\setlength{\tabcolsep}{2.5pt}
\renewcommand{\arraystretch}{1.0}
\caption{Evaluator-level scores (1--5), averaged over 25 cases per evaluator. Acc, Val, Cla, Use, and Avg denote Accuracy, Validity, Clarity, Usefulness, and overall average. The final row gives reported means, which may differ from means of the rounded evaluator rows.}
\label{tab:raters_all}
\resizebox{0.98\linewidth}{!}{%
\begin{tabular}{@{}l|cccc|c|cccc|c|cccc|c|cccc|c@{}}
\toprule
\multirow{2}{*}{Expert}
& \multicolumn{5}{c|}{SAGE}
& \multicolumn{5}{c|}{LLMAD}
& \multicolumn{5}{c|}{TSAD-Agents}
& \multicolumn{5}{c}{TAMA} \\
\cmidrule(lr){2-6} \cmidrule(lr){7-11} \cmidrule(lr){12-16} \cmidrule(lr){17-21}
& Acc & Val & Cla & Use & Avg
& Acc & Val & Cla & Use & Avg
& Acc & Val & Cla & Use & Avg
& Acc & Val & Cla & Use & Avg \\
\midrule
E1  & 4.20 & 3.76 & 3.68 & 3.52 & 3.79 & 3.92 & 3.48 & 3.56 & 3.12 & 3.52 & 3.12 & 2.88 & 2.96 & 2.56 & 2.88 & 2.48 & 2.36 & 2.48 & 1.92 & 2.31 \\
E2  & 4.16 & 4.08 & 4.16 & 4.28 & 4.17 & 3.80 & 3.20 & 3.72 & 2.84 & 3.39 & 2.48 & 2.96 & 3.36 & 2.56 & 2.84 & 2.08 & 2.76 & 2.92 & 2.08 & 2.46 \\
E3  & 3.96 & 4.20 & 3.88 & 3.96 & 4.00 & 3.68 & 3.80 & 3.68 & 3.32 & 3.62 & 3.04 & 3.08 & 3.44 & 2.88 & 3.11 & 2.12 & 2.36 & 2.40 & 2.20 & 2.27 \\
E4  & 4.16 & 4.16 & 3.96 & 3.96 & 4.06 & 4.00 & 4.12 & 4.12 & 4.00 & 4.06 & 2.52 & 2.72 & 2.48 & 2.56 & 2.57 & 1.88 & 2.04 & 2.08 & 1.88 & 1.97 \\
E5  & 4.24 & 4.32 & 4.44 & 4.36 & 4.34 & 3.80 & 3.76 & 4.00 & 3.40 & 3.74 & 3.16 & 3.32 & 3.28 & 2.76 & 3.13 & 2.32 & 3.08 & 3.04 & 2.48 & 2.73 \\
E6  & 3.56 & 3.76 & 3.76 & 3.48 & 3.64 & 3.20 & 3.00 & 2.68 & 2.52 & 2.85 & 2.80 & 2.64 & 2.40 & 2.08 & 2.48 & 1.88 & 2.44 & 2.24 & 1.76 & 2.08 \\
E7  & 4.52 & 4.80 & 4.84 & 4.76 & 4.73 & 4.04 & 4.12 & 3.92 & 3.84 & 3.98 & 3.16 & 3.48 & 3.64 & 3.08 & 3.34 & 2.28 & 2.48 & 3.20 & 2.12 & 2.52 \\
E8  & 4.32 & 4.08 & 4.44 & 4.32 & 4.29 & 3.76 & 3.20 & 3.28 & 2.76 & 3.25 & 3.12 & 3.12 & 3.00 & 2.52 & 2.94 & 2.44 & 2.60 & 2.68 & 2.04 & 2.44 \\
E9  & 4.32 & 4.52 & 4.76 & 4.76 & 4.59 & 4.00 & 3.48 & 2.40 & 1.88 & 2.94 & 2.84 & 2.76 & 2.04 & 1.76 & 2.35 & 2.12 & 2.32 & 2.20 & 1.60 & 2.06 \\
E10 & 4.24 & 4.16 & 4.20 & 4.36 & 4.24 & 3.52 & 3.36 & 3.36 & 3.00 & 3.31 & 2.64 & 2.24 & 2.92 & 2.36 & 2.54 & 2.52 & 2.24 & 3.20 & 2.36 & 2.58 \\
\midrule
\textbf{Mean} & \textbf{4.17} & \textbf{4.18} & \textbf{4.21} & \textbf{4.18} & \textbf{4.18} & 3.77 & 3.55 & 3.47 & 3.07 & 3.47 & 2.89 & 2.92 & 2.95 & 2.51 & 2.82 & 2.21 & 2.47 & 2.64 & 2.04 & 2.34 \\
\bottomrule
\end{tabular}%
}
\end{table}

\end{landscape}

\subsection{Reported statistical comparisons}
The signed-rank tests use $n=250$ evaluator--sample aggregate pairs. Repeated evaluators and samples induce dependence, so the reported $p$-values do not establish significance under a model accounting for both sources of clustering. Evaluator-level summaries provide descriptive robustness information. These summaries are descriptive evidence, not a correction for the dependence in the original tests.

\begin{table}[H]
\centering
\footnotesize
\caption{Wilcoxon signed-rank test results for pairwise comparisons against SAGE.}
\label{tab:ch5_human_eval_stats}
\begin{tabular}{@{}lccc@{}}
\toprule
Comparison & \(n\) & Test statistic & \(p\)-value \\
\midrule
SAGE vs. LLMAD & 250 & 3750 & \(< 0.001\) \\
SAGE vs. TSAD-Agents & 250 & 790 & \(< 0.001\) \\
SAGE vs. TAMA & 250 & 107 & \(< 0.001\) \\
\bottomrule
\end{tabular}
\end{table}

\subsection{Scoring anchors and evaluation interface}
\begin{figure}[H]
\centering
\includegraphics[width=\linewidth]{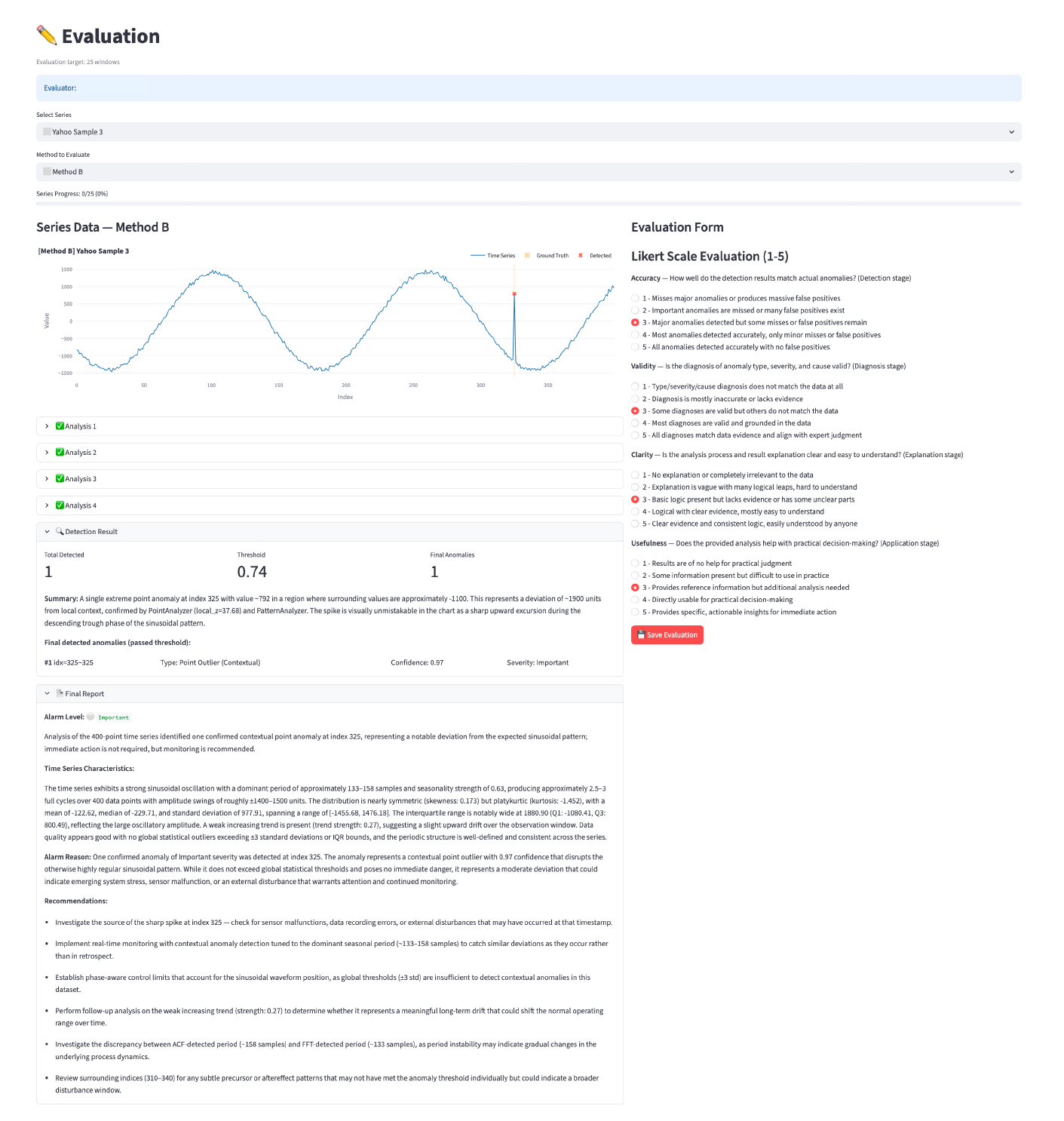}
\caption{Method-blind evaluation interface showing the signal, anomaly intervals, explanation, and scoring controls.}
\label{fig:human_interface}
\end{figure}

The rubric separates detection correctness (Accuracy) from whether a diagnosis fits the observed evidence (Validity), how clearly it is explained (Clarity), and its perceived practical value (Usefulness). Evaluators saw both true and predicted intervals, so these ratings assess the combined detection and explanation output rather than explanation text in isolation. Table~\ref{tab:ch5_human_eval_rubric} gives the score anchors, and Figure~\ref{fig:human_interface} shows the displayed evidence and rating controls.
\begin{table}[H]
\centering
\footnotesize
\setlength{\tabcolsep}{3pt}
\renewcommand{\arraystretch}{0.88}
\caption{Likert-scale criteria and anchor descriptions used in the human evaluation.}
\label{tab:ch5_human_eval_rubric}
\begin{tabular}{@{}p{0.13\textwidth}p{0.07\textwidth}p{0.70\textwidth}@{}}
\toprule
Criterion & Score & Anchor description \\
\midrule
\multirow{5}{*}{Accuracy}
& 1 & Major anomalies are missed, or many normal regions are falsely identified as anomalous. \\
& 2 & Important anomalies are missed, or substantial false positives are present. \\
& 3 & Major anomalies are detected, but some misses or false positives remain. \\
& 4 & Most anomalies are correctly detected, with only minor misses or false positives. \\
& 5 & All anomalies are correctly detected with no false positives. \\
\midrule
\multirow{5}{*}{Validity}
& 1 & The diagnosed anomaly type, severity, or explanation is entirely inconsistent with the observed data. \\
& 2 & The diagnosis is largely inaccurate or weakly supported by the data. \\
& 3 & Some parts of the diagnosis are plausible, but others are not well supported by the data. \\
& 4 & Most diagnostic conclusions are plausible and grounded in the observed evidence. \\
& 5 & All diagnostic conclusions are well supported by the data and closely align with expert judgment. \\
\midrule
\multirow{5}{*}{Clarity}
& 1 & The explanation is absent, unintelligible, or unrelated to the data. \\
& 2 & The explanation is vague and contains substantial logical gaps, making it difficult to follow. \\
& 3 & The overall logic is understandable, but some parts are weakly justified or difficult to interpret. \\
& 4 & The explanation is mostly clear, logically structured, and supported by relevant evidence. \\
& 5 & The explanation is highly clear, coherent, and easy to understand, with strong evidence support throughout. \\
\midrule
\multirow{5}{*}{Usefulness}
& 1 & The analysis is not useful for practical decision-making. \\
& 2 & Some information is provided, but practical use remains limited. \\
& 3 & The analysis provides partially useful information, but additional interpretation is needed before action can be taken. \\
& 4 & The analysis is directly useful for practical decision-making. \\
& 5 & The analysis provides concrete and actionable insights that can support immediate operational response. \\
\bottomrule
\end{tabular}
\end{table}

\FloatBarrier
\section{Diagnostic Case Studies}
\label{app:cases}
\begin{landscape}
\begin{figure}[H]
\centering
\includegraphics[width=0.98\linewidth]{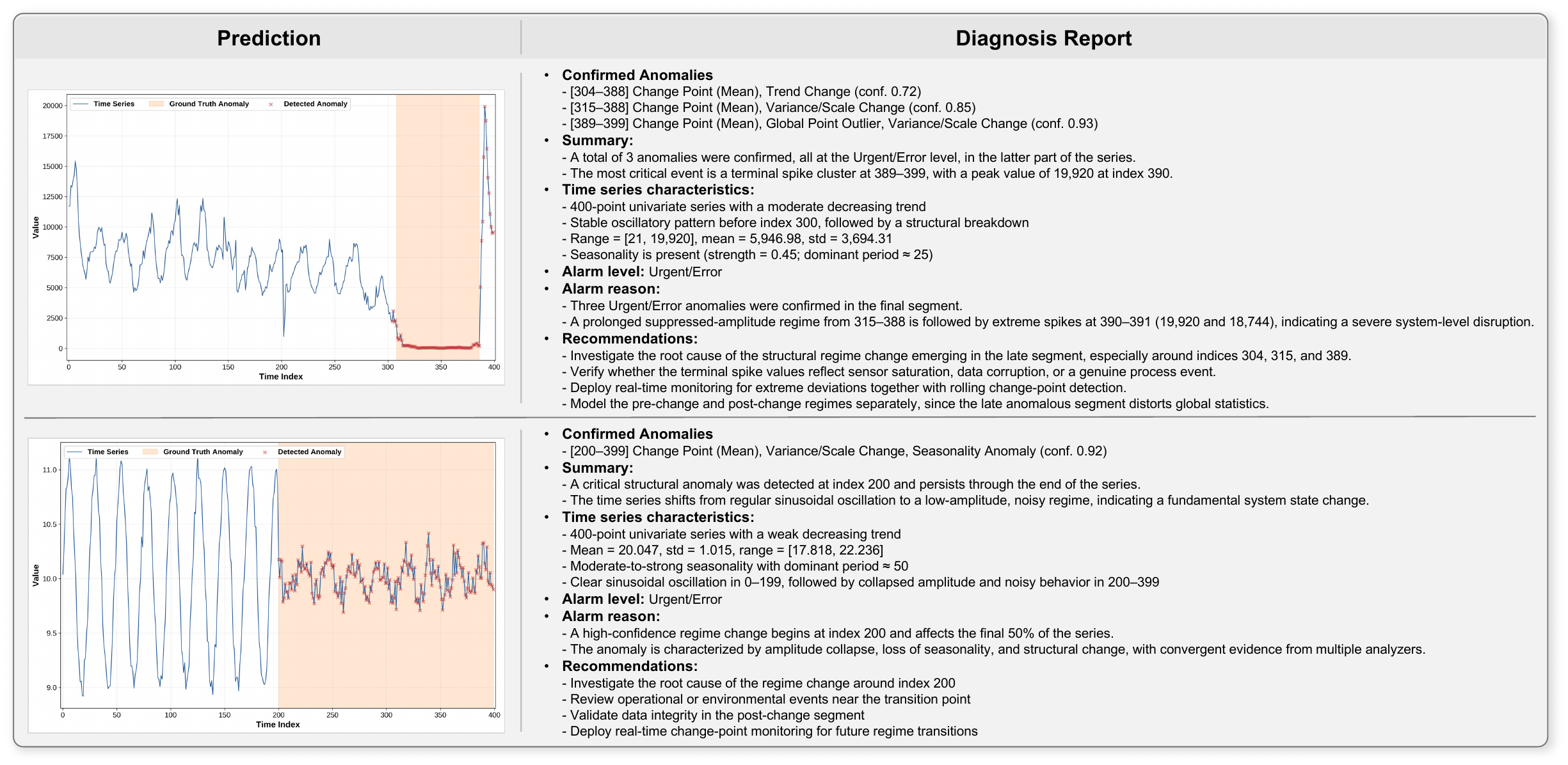}
\caption{Complete SAGE case-study figure. Each row pairs the input series and detected intervals (left) with the generated diagnosis (right). Suggested follow-up checks are investigation prompts, not verified causal explanations.}
\label{fig:cases}
\end{figure}
\end{landscape}

Figure~\ref{fig:cases} pairs each 400-point signal with SAGE's confirmed intervals and analyst-facing report. Shading marks the reference anomaly interval; red crosses mark detected anomalous points. The report records the interval, candidate types, evidence-strength confidence, alarm level, and suggested follow-up checks. These examples illustrate the form of the diagnosis rather than estimate detection accuracy.

In the upper case, the oscillatory signal loses amplitude after approximately index 304 and ends with extreme spikes. SAGE reports overlapping structural changes over 304--388 and 315--388, then a separate spike interval at 389--399 with a global point-outlier label. In the lower case, regular oscillation over 0--199 collapses into a low-amplitude noisy regime; SAGE assigns one extended interval, 200--399, with structural and seasonal labels. The suggested checks, such as inspecting sensor integrity or events near a change point, are hypotheses for an analyst to investigate, not verified root causes.

\end{document}